\documentclass[10pt,twocolumn,letterpaper]{article}

\usepackage[pagenumbers]{cvpr} 

\usepackage{url}
\usepackage{booktabs}
\usepackage[table]{xcolor}
\usepackage[most]{tcolorbox}
\usepackage{graphicx}
\usepackage{multirow}
\usepackage{enumitem}
\usepackage{array}
\usepackage[normalem]{ulem}
\usepackage{float}
\usepackage{tabularx}
\DeclareRobustCommand{\graytext}[1]{%
  \begingroup
  \setlength{\fboxsep}{1.5pt}%
  \colorbox{gray!20}{#1}%
  \endgroup
}
\newtcolorbox{promptbox}[1][]{
    enhanced,
    colback=blue!4,
    colframe=blue!55!black,
    coltitle=white,
    fonttitle=\bfseries,
    title={#1},
}
\newtcolorbox{orangepromptbox}[1][]{
    enhanced,
    colback=orange!4,
    colframe=orange!55!black,
    coltitle=white,
    fonttitle=\bfseries,
    title={#1},
}
\newcolumntype{C}{>{\centering\arraybackslash}X}

\definecolor{cvprblue}{rgb}{0.21,0.49,0.74}
\usepackage[pagebackref,breaklinks,colorlinks,allcolors=cvprblue]{hyperref}

\def\confName{CVPR}
\def\confYear{2027}

\title{FG-GraphBench: Benchmarking and Enhancing Fine-Grained Visual Understanding with Multimodal Scene Graphs}

\author{Sicheng Liu$^{1}$, Xunkai Li$^{1}$, Ziyu Han$^{2}$, Xingnan Zhang$^{1}$, Ru Zhang$^{1}$, Rong-Hua Li$^{1,*}$\\[0.5em]
$^{1}$Department of Computer Science, Beijing Institute of Technology\\
$^{2}$School of Airspace Science and Engineering, Shandong University\\
}

\begin{document}
\maketitle
\begingroup
\renewcommand\thefootnote{\fnsymbol{footnote}}
\setcounter{footnote}{1} 
\footnotetext{Corresponding author. Email: lironghuabit@126.com}
\endgroup
\begin{abstract}
Fine-grained visual understanding requires multimodal large language models (MLLMs) to recognize small visual entities and understand their relationships. 
Scene graphs offer a natural way to represent this information, organizing visual entities as nodes and their relationships as edges.
However, we identify two challenges: existing scene graph benchmarks provide insufficient coverage of fine-grained entities and the relationships among them, and existing methods for fine-grained visual understanding have yet to effectively leverage such graph-structured information in MLLMs.
To address these challenges, we introduce Fine-Grained Multimodal Scene Graph Benchmark (FG-GraphBench) and FiGRA. 
FG-GraphBench is a benchmark comprising three complementary tasks: fine-grained visual question answering (FG-QA), image captioning (FG-Caption), and image description generation (FG-Description). 
We develop a two-stage construction pipeline that discovers and grounds fine-grained entities and establishes their relations to construct the fine-grained multimodal scene graphs (MSGs) in FG-GraphBench.
Building on these MSGs, we propose FiGRA, a \textbf{\underline{Fi}}ne-grained \textbf{\underline{G}}raph \textbf{\underline{R}}epresentation \textbf{\underline{A}}ugmentation method for MLLMs. 
FiGRA adopts a two-stage training strategy.
It encodes multimodal entity and relation features as continuous graph tokens, and integrates them into the MLLM. 
Experiments show that FiGRA significantly outperforms strong baselines across three tasks of FG-GraphBench. 
Further analyses confirm the effectiveness of FG-GraphBench and highlight the importance of preserving fine-grained entity and relation information for visual understanding.
\end{abstract}

\section{Introduction}
Recent advances in multimodal large language models (MLLMs) have led to remarkable progress in a wide range of vision-language tasks~\citep{liu2023visual,dai2023instructblip,liu2024improved,wang2024qwen2}. 
However, many real-world tasks require fine-grained visual understanding. 
Current MLLMs often struggle to recognize small objects and understand the relationships among them~\citep{zhong2026focus,shen2025zoomeye,wu2024v,chen2024spatialvlm,lian2025describe}. 
As a result, they may overlook visual details and produce hallucinations~\citep{leng2024mitigating}.

A promising way to address these limitations is to introduce explicit structural priors into visual understanding. 
Scene graphs organize visual entities as nodes and their relationships as edges~\citep{hudson2019gqa,krishna2017visual,yang2022panoptic}.
Their nodes provide direct \textbf{object-level} cues for recognizing fine-grained objects, while their edges provide explicit \textbf{relation-level} cues that MLLMs would otherwise need to infer from image~\citep{shi2019explainable}.

However, realizing these benefits in MLLMs presents two key challenges.
\textit{\textbf{\underline{Challenge 1:} How can we obtain scene graphs that are sufficiently fine-grained to provide the object-level and relation-level cues needed for fine-grained visual understanding?}}
Existing scene-graph-based benchmarks were largely developed for conventional scene understanding and typically emphasize relatively salient objects, while fine-grained entities and their relationships remain insufficiently covered~\cite{krishna2017visual,hudson2019gqa}.
Moreover, these benchmarks are not specifically designed to evaluate fine-grained visual understanding~\cite{zhuang2018hcvrd,grunde2021agqa,qian2024nuscenes}.

\textit{\textbf{\underline{Challenge 2:} How can we effectively incorporate such fine-grained graph information into MLLMs so that it enhances both fine-grained object perception and relation understanding?}}
Several recent approaches to fine-grained visual understanding improve access to local visual evidence through visual cropping~\citep{khayatkhoei2025mllms,zhong2026focus}, hierarchical zooming~\citep{shen2025zoomeye}, patch-wise image decomposition~\citep{peng2025matters}, or high-resolution encoding~\citep{lian2025describe,hua2025finecaption}.
While these strategies can enhance the perception of small or local visual details, they generally treat fine-grained information as visual regions or patches rather than explicitly modeling the structured dependencies among entities.
Conversely, prior graph-based approaches have explored structured representations for visual reasoning~\cite{wang2023vqa,shi2019explainable}, but were not designed to support fine-grained visual understanding in MLLMs.

To address \textit{\textbf{\underline{Challenge 1}}}, we introduce FG-GraphBench, a benchmark for fine-grained visual understanding with fine-grained MSGs. 
FG-GraphBench comprises three complementary tasks: FG-QA for fine-grained visual question answering, FG-Caption for fine-grained image captioning, and FG-Description for fine-grained image description generation. 
Together, these tasks assess MLLMs' fine-grained entity recognition and relational reasoning capabilities.
In particular, FG-QA places greater emphasis on relational reasoning, whereas FG-Caption and FG-Description require models to recognize fine-grained entities and capture their relationships in generated text.
We construct FG-GraphBench through a two-stage pipeline. 
Stage 1 enriches the source entity set through vocabulary-based grounding to localize source entities and detect additional instances, followed by two rounds of discovery using a vision-language model (VLM) to identify new fine-grained entities, with two-stage filtering to remove unreliable entities.
Stage 2 predicts holistic visual relations and constructs dependency relations using a dependency candidate tree, integrating the resulting relations with the enriched entity set to form the final fine-grained MSGs.

To address \textit{\textbf{\underline{Challenge 2}}}, we propose \textbf{FiGRA}, a \textbf{\underline{Fi}}ne-grained \textbf{\underline{G}}raph \textbf{\underline{R}}epresentation \textbf{\underline{A}}ugmentation method for MLLMs.
It encodes MSGs into continuous graph tokens and injects them into the MLLM alongside visual inputs. 
FiGRA first performs multimodal graph encoding to obtain entity and relation representations, then projects them into continuous graph tokens and integrates them into the MLLM.
It adopts a two-stage training strategy, consisting of a pre-training stage with a frozen MLLM and a supervised fine-tuning stage for downstream tasks.

In summary, our key contributions are as follows:
\textbf{\textit{(1) New Perspective.}}
To the best of our knowledge, we are the first to explore fine-grained MSGs as a unified structural representation for enhancing fine-grained visual understanding in MLLMs.
\textbf{\textit{(2) New Benchmark and Data Construction Pipeline.}}
We develop a two-stage pipeline to construct fine-grained MSGs and build FG-GraphBench, comprising FG-QA, FG-Caption, and FG-Description.
\textbf{\textit{(3) New Approach.}}
We propose FiGRA, which encodes multimodal entity and relation features into continuous graph tokens and integrates them into MLLMs through two-stage training.
\textbf{\textit{(4) SOTA Performance.}}
FiGRA achieves the best results on the primary evaluation metrics among evaluated baselines across all three tasks.

\section{Related works}
\textbf{Fine-Grained Visual Understanding in MLLMs.}
Recent work improves fine-grained visual understanding in MLLMs through high-resolution image encoding and adaptive visual search, enabling models to access local visual details more effectively~\citep{li2024monkey,guo2024llava,zhong2026focus,shen2025zoomeye}. Beyond visual perception, region-level understanding has been explored through referring, grounding, segmentation, and open-world visual recognition~\citep{guo2024regiongpt,you2024ferret,rasheed2024glamm,lai2024lisa,zhang2024omg,wang2024all}. 
For image description, existing methods enhance local visual details or generate region-level captions at different granularities~\citep{peng2025matters,hua2025finecaption,lian2025describe}. 

\textbf{Scene Graphs for Visual Understanding.}
Scene graphs represent objects, attributes, and their relations~\cite{krishna2017visual,hudson2019gqa}. 
Existing fine-grained scene graph generation methods distinguish semantically similar predicates and expand predicate vocabularies~\cite{lyu2022fine,zhang2022fine,yu2023visually}. 
There are also studies that address challenges in relation learning and annotation quality in scene graph generation~\cite{min2023environment,kim2024adaptive,zhu2024local}.
Beyond scene graph generation, scene graphs have also been used for explicit visual reasoning~\cite{shi2019explainable,wang2023vqa,zhang2021explicit}.

\section{Data construction pipeline and FG-GraphBench}
\label{sec:mgqa_benchmark}

\subsection{Benchmark overview}
\label{sec:benchmark_overview}

\begin{figure*}[!t]
    \centering
    \includegraphics[width=\textwidth]{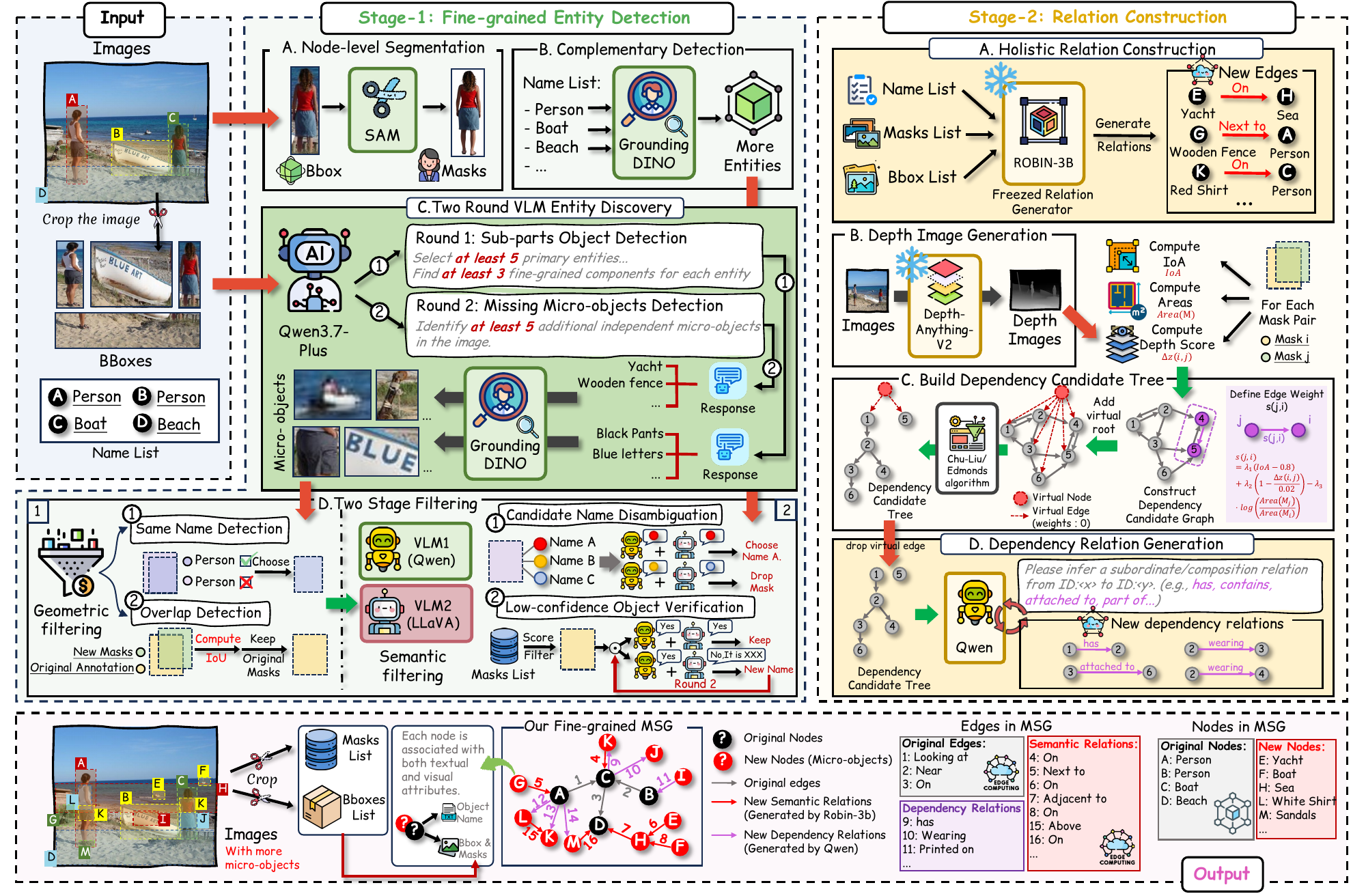}
    \caption{Overview of the FG-GraphBench construction pipeline.}
    \label{fig:data_pipeline}
    \vspace{-3mm}
\end{figure*}

Existing scene graph benchmarks provide insufficient annotations of fine-grained objects and their relationships. To address this limitation, we develop a unified data construction pipeline for discovering fine-grained objects and constructing their relationships, and use this pipeline to build FG-GraphBench, a fine-grained multimodal scene-graph benchmark.

As illustrated in~\cref{fig:data_pipeline}, our construction pipeline consists of two stages: \textbf{Stage~1} discovers and verifies fine-grained objects, while \textbf{Stage~2} constructs their relationships.


\subsection{Stage 1: fine-grained entity detection and verification}
\label{sec:object_discovery}

Stage~1 constructs a unified entity set that covers both small independent objects and fine-grained entities associated with larger objects. 
We begin with a scene graph, which provides source entities (with bounding boxes) and annotated relations. 
Complete implementation details are provided in Sec. B of Supplementary Material.

\textbf{Initial vocabulary-based grounding.}
We first retain source entities and use their bounding boxes with SAM~\citep{kirillov2023segment} to obtain pixel-level masks. We then query GroundingDINO~\citep{liu2024grounding} with source entity names to detect additional instances matching these names, and segment the resulting regions with SAM. 

\textbf{Two Round VLM Fine-grained Entity Discovery.}
Inspired by~\citet{ramanathan2023paco} and~\citet{pan2023towards}, we consider fine-grained visual entities to manifest in two forms: standalone entities and object-dependent entities. 
The former refer to independent fine-grained entities in the scene. 
The latter are visually associated with larger entities via dependency relations (e.g., part of and attached to).
This distinction motivates a two-round discovery strategy designed to detect the two forms of entities separately.
In the first round, the VLM (Qwen3.7-Plus) receives the image and a simplified source scene graph, selects source entities as visual anchors, and searches around them for fine-grained components. 
In the second round, it additionally takes the first-round discoveries as context and rescans the full image for standalone entities. 
The VLM outputs are subsequently used for grounding with GroundingDINO and segmentation with SAM, followed by geometric and semantic verification before inclusion in the final entity set.

\textbf{Two-stage filtering.}
We merge candidates from vocabulary-based grounding and two-round VLM discovery.
We apply two filtering stages. 
First, we apply geometric filtering to merge overlapping candidates with the same name and remove spatially inconsistent candidate, while preserving the original source entities.
Second, we use Qwen2-VL and LLaVA for semantic filtering.
For highly overlapping candidates with different names, the models select the name that best matches the region. 
We also examine candidates with low grounding confidence to determine whether the current name matches the corresponding mask.
If the two models do not both confirm the current name, a new name is generated. We then run GroundingDINO and SAM again and keep the new name only when the resulting mask is consistent with the original region (see Sec. B.3).

\subsection{Stage 2: Relation construction}
\label{sec:relation_construction}

Stage~1 expands the source entity set with newly discovered fine-grained entities, and Stage~2 infers relations involving them.
These include general visual relations (e.g., \emph{next to} and \emph{riding}) that capture relationships among entities in the broader scene, and dependency relations (e.g., a button \emph{on} a shirt) that characterize how fine-grained entities are associated with larger objects.
We therefore construct relations through two pathways: \textbf{Holistic Relation Construction} and \textbf{Dependency Relation Construction with the Dependency Candidate Tree}.
See Sec. C for details.

\subsubsection{Holistic relation construction}
\label{sec:holistic_relation}

ROBIN~\citep{park2025synthetic} is a multimodal language model trained for dense scene graph generation, capable of predicting visual relationships among specified entities. 
We employ ROBIN to construct relations across all entities. 
Given the image and all grounded entity regions, ROBIN predicts their relations in the context of the full scene.
Since relations between original entities are already provided by the source annotations, we retain only ROBIN predictions involving at least one newly discovered entity.

\subsubsection{Dependency relation construction with the dependency candidate tree}
\label{sec:subordinate_relation}

\textbf{Dependency Candidate Tree Construction.}
We first use Depth Anything V2~\citep{yang2024depth} to estimate a dense depth map for each image.
For each target entity, we identify potential parent entities whose masks substantially cover the target and exhibit consistent local depths across the target's mask boundary.
We score each candidate connection using mask coverage, local depth consistency, and an area-ratio penalty that discourages excessively large parents.
We then add a virtual root with zero-weight connections to all entities and apply the Chu--Liu/Edmonds algorithm~\citep{chu1965shortest,edmonds1967optimum} to obtain a maximum spanning arborescence.
Removing the virtual root yields the \textbf{Dependency Candidate Tree}, in which entities without a suitable parent remain independent.

\textbf{Dependency Relation Prediction.}
For each edge in the Dependency Candidate Tree involving at least one newly discovered entity, we provide Qwen2-VL-7B-Instruct~\citep{wang2024qwen2} with the image and the names of the two connected entities.
The model predicts a dependency relation (e.g., has, contains) to form a directed relation triplet.
These dependency relations are integrated with the original relations and holistic relation predictions to construct the final fine-grained MSG (see Sec. C.2).

\subsection{FG-GraphBench}
\label{sec:benchmark_tasks}

Using the above pipeline, we construct FG-GraphBench with three datasets for complementary tasks.
Built from GQA dataset~\citep{hudson2019gqa}, FG-QA evaluates fine-grained relational question answering.
Built from Action Genome~\citep{ji2020action}, FG-Description evaluates detailed image description.
Built from MS COCO Captions dataset~\citep{chen2015microsoft}, FG-Caption evaluates concise image caption generation with fine-grained details.
Each sample in FG-GraphBench consists of an image, its corresponding MSG, a question, and the associated text response.
We use Qwen3.7-Plus to construct fine-grained questions and corresponding responses (see Sec. A.7).

Within FG-QA, we categorize questions into four types: \textbf{queryRel} for identifying an entity through its relation to another entity,
\textbf{verifyRel} for determining whether a specified relation holds, 
\textbf{existRel} for asking whether an entity satisfying given relational constraints exists, 
and \textbf{chooseRel} for selecting the correct relation from candidate options. 
\Cref{tab:benchmark_statistics} summarizes statistics for FG-GraphBench.

We further conduct a data quality check to assess the reliability of the constructed MSGs. The results demonstrate high annotation accuracy, achieving 88.0\%--90.4\% entity accuracy and 83.4\%--86.0\% relation accuracy across the three datasets (see Sec. A.6).
More detailed information about the datasets is provided in Sec. A.


\begin{table}[t]
    \centering
    \caption{Statistics of FG-GraphBench. We report the number of graphs and the average numbers of nodes and edges per graph.}
    \label{tab:benchmark_statistics}
    \small
    \setlength{\tabcolsep}{3pt}
    \vspace{-2mm}
    \begin{tabularx}{\columnwidth}{@{}Xrrr@{}}
        \toprule
        Dataset / Question Type
        & Graphs & Avg. Nodes & Avg. Edges \\
        \midrule
        FG-Description (AG)
        & 9,595 & 16.19 & 35.08 \\
        FG-Caption
        & 14,026 & 35.05 & 101.32 \\
        \midrule
        FG-QA (Overall)
        & 43,384 & 28.67 & 81.22 \\
        \quad queryRel
        & 24,824 & 28.93 & 81.52 \\
        \quad verifyRel
        & 6,977 & 28.23 & 80.04 \\
        \quad existRel
        & 6,974 & 28.15 & 80.24 \\
        \quad chooseRel
        & 4,609 & 28.79 & 82.84 \\
        \bottomrule
    \end{tabularx}
    \vspace{-3mm}
\end{table}

\section{FiGRA: fine-grained graph representation augmentation method for multimodal large language models}
\label{sec:figra}

\begin{figure*}[!t]
    \centering
    \includegraphics[width=\textwidth]{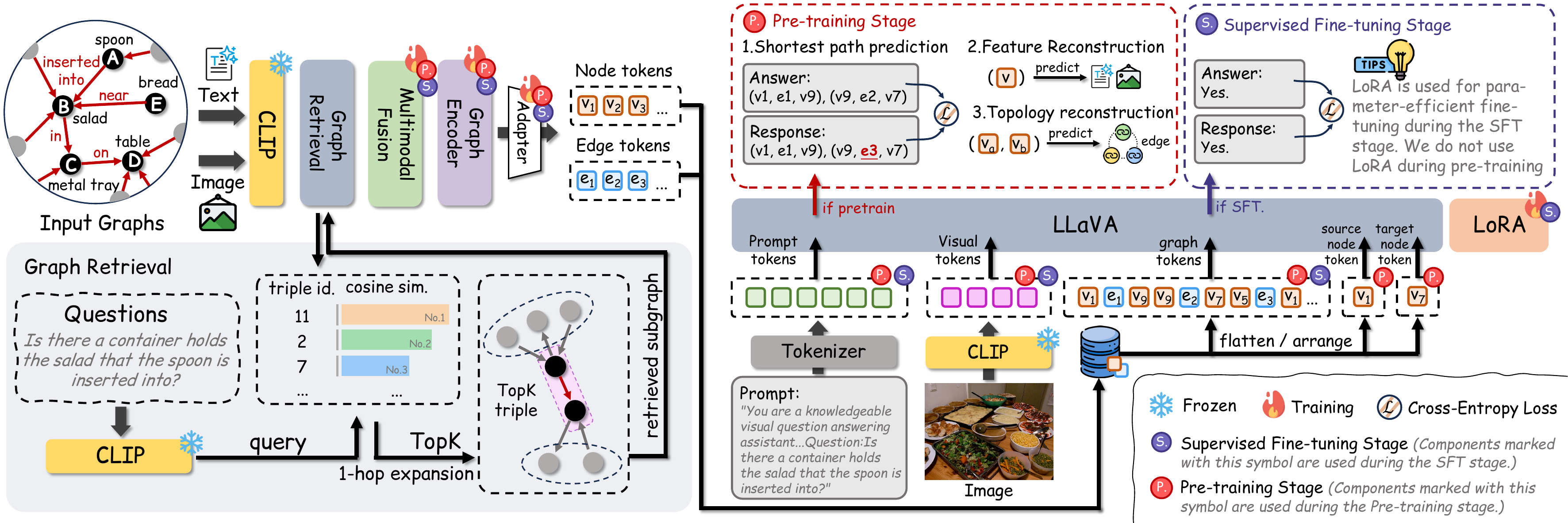}
    \caption{Overview of FiGRA and its two-stage training strategy.}
    \label{fig:figra_framework}
    \vspace{-3mm}
\end{figure*}

In this section, we introduce \textbf{FiGRA}: a \textbf{\underline{Fi}}ne-grained \textbf{\underline{G}}raph \textbf{\underline{R}}epresentation \textbf{\underline{A}}ugmentation method for MLLMs.
It leverages our fine-grained MSGs to enhance MLLMs' understanding of fine-grained visual objects and their relationships, as illustrated in~\cref{fig:figra_framework}. 

\subsection{Multimodal graph encoding}
\label{sec:multimodal_graph_encoding}
Given a fine-grained MSG $\mathcal{G}=(\mathcal{V},\mathcal{E})$, we first use the pretrained CLIP ViT-L/14@336px~\citep{radford2021learning} to extract node text features $\mathbf{x}_v^{text}$ and visual features $\mathbf{x}_v^{vision}$ from its corresponding image region.
For edge from node $u$ to node $v$, we encode the relation text to obtain the edge feature $\mathbf{x}_{uv}$.

\textbf{Question-Relevant Graph Retrieval.}
For FG-QA and FG-Caption, we encode the question with the same CLIP text encoder to obtain the query embedding. 
We represent each triple $(u,r,v)$ by averaging its head and tail node features with its relation feature:
\begin{equation}
\mathbf{h}_{(u,r,v)}^{m}
=\frac{1}{3}\left(
\mathbf{x}_u^{m}+\mathbf{x}_{uv}+\mathbf{x}_v^{m}
\right),
\end{equation}
where $m=text$ for FG-QA and $m=vision$ for FG-Caption. (For FG-Caption, we use visual node features because image captioning aims to summarize an image's primary visual content.)
We rank triple representations by cosine similarity to the query embedding and retrieve the top-$K$ triples $\mathcal{E}{\mathrm{top}}$. 
The nodes in these triples form an initial node set, and we use 1-hop expansion to build retrieved subgraph $\mathcal{E}_q$:
\vspace{-3mm}
\begin{equation}
\begin{aligned}
\mathcal{V}_{\mathrm{init}}
&=\bigcup_{(u,r,v)\in\mathcal{E}_{\mathrm{top}}}\{u,v\},\\
\mathcal{E}_q
&=\{(a,r,b)\in\mathcal{E}\mid
a\in\mathcal{V}_{\mathrm{init}}
\lor b\in\mathcal{V}_{\mathrm{init}}\}.
\end{aligned}
\end{equation}
The retrieved subgraph is $\mathcal{G}_q=(\mathcal{V}_q,\mathcal{E}_q)$, where $\mathcal{V}_q=\{a,b\mid(a,r,b)\in\mathcal{E}_q\}$.
For FG-Description, we use the complete graph.

\textbf{Multimodal Fusion and Graph Encoding.}
We project each node's textual and visual features into a shared space, obtaining $\mathbf{z}_v^{text}\in\mathbb{R}^{d}$ and $\mathbf{z}_v^{vision}\in\mathbb{R}^{d}$.
We then adaptively combine them through a dimension-wise gate:
\begin{equation}
\begin{aligned}
\mathbf{g}_v
&=\sigma\!\left(W_g[\mathbf{z}_v^{text};\mathbf{z}_v^{vision}]+b_g\right),\\
\mathbf{h}_v^{(0)}
&=\mathbf{g}_v\odot\mathbf{z}_v^{text}
 +(1-\mathbf{g}_v)\odot\mathbf{z}_v^{vision},
\end{aligned}
\end{equation}
Here, $\mathbf{g}_v$ is a dimension-wise gate that balances textual and visual information.
Edge features are independently projected as $\mathbf{e}_{uv}=\phi_{edge}(\mathbf{x}_{uv})$.
Node features capture individual entities but do not explicitly model their interactions.
We therefore use an edge-aware Graph Transformer encoder~\cite{dwivedi2020generalization} to propagate relational information by aggregating neighboring node features and their corresponding relation embeddings:
\begin{equation}
\mathbf{h}_v^{(\ell+1)}
=\operatorname{GT}_{\ell}\!\left(
\mathbf{h}_v^{(\ell)},
\left\{
(\mathbf{h}_u^{(\ell)},\mathbf{e}_{uv})
\mid (u,r,v)\in\mathcal{E}_{q}
\right\}
\right).
\end{equation}
Here, $\mathbf{h}_v^{(\ell)}$ denotes the representation of node $v$ at layer $\ell$, $\operatorname{GT}_{\ell}$ denotes the corresponding Graph Transformer layer.

\subsection{Graph token integration with MLLMs}
\label{sec:graph_token_integration}
We first project the final node and edge representations into the MLLM's hidden space using a multilayer perceptron (MLP) with two layers:
\begin{equation}
\begin{aligned}
\tilde{\mathbf{h}}_v
&=\operatorname{GELU}(\mathbf{h}_v^{(L)}W_1+b_1)W_2+b_2,\\
\tilde{\mathbf{e}}_{uv}
&=\operatorname{GELU}(\mathbf{e}_{uv}W_1+b_1)W_2+b_2.
\end{aligned}
\end{equation}

We then construct the graph token sequence.
Each directed edge $(u,r,v)\in\mathcal{E}_{q}$ contributes a triplet $[\tilde{\mathbf{h}}_u,\tilde{\mathbf{e}}_{uv},\tilde{\mathbf{h}}_v]$. 
We concatenate these triplets in the original edge order, followed by the representations of isolated nodes:
\begin{equation}
\mathbf{T}_g
=
\mathop{\Vert}_{(u,r,v)\in\mathcal{E}_{q}}
[\tilde{\mathbf{h}}_u,\tilde{\mathbf{e}}_{uv},\tilde{\mathbf{h}}_v]
\;\Vert\;
\mathop{\Vert}_{w\in\mathcal{V}_{\mathrm{iso}}}
[\tilde{\mathbf{h}}_w],
\end{equation}
where $\Vert$ denotes sequence concatenation and $\mathcal{V}_{\mathrm{iso}}$ contains nodes without incident edges in the selected graph.
Thus, $\mathbf{T}_g\in\mathbb{R}^{N_g\times d_{\mathrm{MLLM}}}$, where $N_g=3|\mathcal{E}_{q}|+|\mathcal{V}_{\mathrm{iso}}|$ is the sequence length and $d_{\mathrm{MLLM}}$ is the MLLM hidden dimension.

We use LLaVA-1.5-7B~\citep{liu2024improved} as the MLLM backbone and insert visual tokens $\mathbf{T}_I$ and graph tokens $\mathbf{T}_g$ at designated positions in the instruction $Q$ (see Sec. D.2 for details). The model generates the output $Y$ autoregressively:
\begin{equation}
p_{\theta}(Y\mid I,\mathcal{G},Q)
=
\prod_{t=1}^{|Y|}
p_{\theta}(y_t\mid y_{<t},\mathbf{T}_I,\mathbf{T}_g,Q).
\end{equation}
where $y_{<t}$ denotes previously generated tokens.

\subsection{Two-stage training}
\label{sec:two_stage_training}

\subsubsection{Pre-training stage}
To enable the MLLM to interpret graph tokens as structured relational information, we pretrain the graph encoder and auxiliary modules using three objectives while keeping the MLLM frozen.
We sample pre-training graphs from the training splits of FG-Description, FG-Caption, and FG-QA in a 4:1:1 ratio.
See Sec. D.1 for details.

\textbf{Shortest-Path Prediction.}
Given a sampled node pair, the MLLM predicts the shortest directed path as a sequence of relation triples, supervised by the autoregressive language modeling loss $\mathcal{L}_{\mathrm{sp}}$.

\textbf{Node Feature Reconstruction.}
Two MLP decoders reconstruct the original textual and visual node features from $\mathbf{H}$ using mean squared error (MSE) losses $\mathcal{L}_{\mathrm{feat}}$.

\textbf{Topology Reconstruction.}
An MLP decoder predicts directed edge existence between node pairs, supervised by binary cross-entropy (BCE) over observed edges and sampled negative pairs ($\mathcal{L}_{\mathrm{topo}}$).

The overall pre-training objective is
\begin{equation}
\mathcal{L}_{\mathrm{pre}}
=\lambda_{\mathrm{sp}}\mathcal{L}_{\mathrm{sp}}
+\lambda_{\mathrm{feat}}\mathcal{L}_{\mathrm{feat}}
+\lambda_{\mathrm{topo}}\mathcal{L}_{\mathrm{topo}}.
\end{equation}

\subsubsection{Supervised fine-tuning stage}
We initialize FiGRA with pretrained weights and fine-tune its trainable components and LoRA~\cite{hu2021lora} parameters on downstream tasks.
For a task instruction $Q$ and reference answer $Y$, we minimize the causal language modeling loss:
\begin{equation}
\mathcal{L}_{\mathrm{SFT}}
=-\frac{1}{|Y|}\sum_k
\log p_{\theta}\!\left(
y_k\mid y_{<k},\mathbf{T}_I,\mathbf{T}_g,Q
\right).
\end{equation}

\begin{table*}[t]
\centering
\scriptsize
\setlength{\tabcolsep}{3.7pt}
\renewcommand{\arraystretch}{1.12}
\caption{Results on the FG-Description.
Methods evaluated without fine-tuning, including training-free methods and some pretrained models used directly for inference, are highlighted in \graytext{gray}.
Best results are highlighted in \textbf{bold}, while second-best results are \underline{underlined}.
}
\vspace{-2mm}
\label{tab:fg_description_results}
\begin{tabularx}{\textwidth}{c l c|*{8}{C}}
\toprule
 & \textbf{Method} & \textbf{LLM} 
 & \textbf{BLEU-1} & \textbf{BLEU-2} & \textbf{BLEU-3} & \textbf{BLEU-4}
 & \textbf{CIDEr} & \textbf{ROUGE-L} & \textbf{METEOR} & \textbf{CLIP-Score} \\
\midrule

\multicolumn{11}{l}{\textit{Open-source Multimodal Large Language Models}} \\

1 & LLaVA-1.5-7B~\cite{liu2024improved}       & 7B  & 0.6754 & 0.4888 & 0.3456 & 0.2494 & 0.1854 & 0.3561 & 0.3992     & 0.5655     \\
2 & ALLaVA-3B~\cite{chen2024allava}              & 3B  & 0.6385 & 0.4363 & 0.2887 & 0.1953 & 0.1911 & 0.5435 & 0.3863     & 0.5898     \\
3 & Llama-3.2-11B-Vision-Instruct~\cite{grattafiori2024llama}    & 11B & 0.6624 & 0.4656 & 0.3215 & 0.2269 & 0.2448 & 0.5691 & 0.4241     & 0.6042     \\
4 & InternVL3.5-8B~\cite{wang2025internvl3}         & 8B  & \underline{0.7203} & \underline{0.5394} & \underline{0.3976} & \underline{0.2985} & \underline{0.2892} & \underline{0.5963} & \underline{0.4542} & 0.6049 \\
5 & Qwen2-VL-7B-Instruct~\cite{wang2024qwen2}   & 7B  
& 0.6969
& 0.5135
& 0.3692
& 0.2694
& 0.2291
& 0.3723
& 0.4207
& 0.6020 \\

\midrule
\multicolumn{11}{l}{\textit{Specialized Methods for Fine-grained Image Captioning}} \\
\rowcolor{gray!20}
6 & Patch Matters~\cite{peng2025matters} & 7B & 0.3408 & 0.1960 & 0.0998 & 0.0511 & 0.0088 & 0.2192 & 0.1217 & 0.5634 \\
\rowcolor{gray!20}
7 & IT~\cite{pi2024image}
& 7B
& 0.5501
& 0.3722
& 0.2344
& 0.1458
& 0.1061
& 0.2823
& 0.3307
& 0.6033 \\
\rowcolor{gray!20}
8 & Synthesized~\cite{dong2024benchmarking}
& 7B
& 0.4281
& 0.2427
& 0.1239
& 0.0645
& 0.0122
& 0.2383
& 0.1800
& \textbf{0.6081} \\
9 & SC-Captioner~\cite{zhang2025sc}   & 7B  
& 0.6455 
& 0.4704 
& 0.3354 
& 0.2435 
& 0.1232 
& 0.5760 
& 0.4011 
& 0.5998 \\

\rowcolor{gray!20}
10 & ShareGPT4V-7B~\cite{chen2024sharegpt4v}  & 7B  
& 0.5185 
& 0.3154 
& 0.1784 
& 0.1035 
& 0.0844 
& 0.4483 
& 0.2693 
& 0.5327 \\

11 & ShareGPT4V-7B~\cite{chen2024sharegpt4v}   & 7B  
& 0.6540 
& 0.4722 
& 0.3382 
& 0.2481 
& 0.2839 
& 0.5680 
& 0.4412 
& \underline{0.6052} \\

\midrule
12 & FiGRA \textit{(ours)}   & 7B  
& \textbf{0.7544} 
& \textbf{0.5914}
& \textbf{0.4593} 
& \textbf{0.3611}
& \textbf{0.4062}
& \textbf{0.5981}
& \textbf{0.4976}
& 0.6037 \\

\bottomrule
\end{tabularx}
\vspace{-3mm}
\end{table*}

\section{Experiments}
\label{sec:experiments}

In this section, we conduct various experiments to verify the effectiveness of FiGRA and FG-GraphBench, focusing on following questions:
\textbf{Q1}: How does FiGRA perform across the three fine-grained visual understanding tasks?
\textbf{Q2}: Do FiGRA's modules and two-stage training strategy contribute to its performance?
\textbf{Q3}: Does FG-GraphBench supply useful fine-grained visual information?
\textbf{Q4}: How do different methods of incorporating graph information into MLLMs affect model performance and computational cost?

\subsection{Experimental settings}
\label{sec:experimental_settings}

\textbf{Baselines.}
We compare FiGRA with two groups of baselines:
\emph{\textbf{(1) Open-source MLLMs}}: including LLaVA-1.5-7B~\cite{liu2024improved}, Qwen2-VL-7B-Instruct~\cite{wang2024qwen2}, ALLaVA-3B~\cite{chen2024allava}, Llama-3.2-11B-Vision-Instruct~\cite{grattafiori2024llama}, and InternVL3.5-8B~\cite{wang2025internvl3}.
\emph{\textbf{(2) Specialized Methods for Fine-Grained Visual Understanding}}: including fine-grained visual question answering (VQA) methods (ViCrop~\citep{khayatkhoei2025mllms}, Zoom Eye~\citep{shen2025zoomeye}, SEAL~\citep{wu2024v}, and PinPoint~\citep{kwon2026focus}), and fine-grained image captioning methods (Patch Matters~\citep{peng2025matters}, IT~\citep{pi2024image}, Synthesized~\citep{dong2024benchmarking}, SC-Captioner~\citep{zhang2025sc}, and ShareGPT4V-7B~\citep{chen2024sharegpt4v}).
We fine-tune open-source MLLMs with LoRA and adapt specialized methods following their respective protocols. Implementation details are provided in Sec. E.

\textbf{Metrics.}
For FG-Description and FG-Caption, we report BLEU scores, CIDEr, ROUGE-L, METEOR, and CLIP-Score. 
We use Entity-Acc and CAPTURE~\citep{dong2024benchmarking} to evaluate fine-grained visual content in generated descriptions.
Entity-Acc measures scene graph entity coverage as the proportion  of unique entity names in each graph appearing in the generated description. 
CAPTURE evaluates the consistency of objects, attributes, and relations extracted from generated descriptions and reference captions. 
For FG-QA, we report overall and per-type accuracy, WUPS~\cite{malinowski2014multi}, and Macro-F1.

\subsection{Overall performance}
\label{sec:overall_performance}
To answer \textbf{Q1}, we compare FiGRA with open-source MLLMs and specialized methods on all three FG-GraphBench tasks, with results reported in \Cref{tab:fg_description_results,tab:fg_qa_results,tab:fg_caption_results}.

As shown in \Cref{tab:fg_description_results,tab:fg_caption_results}, FiGRA achieves consistently strong performance across detailed image description generation and image captioning tasks. 
These results demonstrate FiGRA's ability to capture fine-grained visual details in both concise captions and detailed descriptions.

As shown in \Cref{tab:fg_qa_results}, FiGRA achieves 59.59\% overall accuracy on FG-QA. 
It outperforms the strongest baseline (PinPoint (Qwen)) by 9.32 percentage points.
It also achieves the highest accuracy across all four question types. 
We attribute these gains to question-relevant subgraph retrieval and graph encoding, which provide the MLLM with relevant entities and their visual and semantic relationships.

We also provide a qualitative case study comparing FiGRA with LLaVA-1.5-7B on FG-GraphBench in Sec. F.

\begin{table*}[t]
\centering
\scriptsize
\setlength{\tabcolsep}{3.2pt}
\renewcommand{\arraystretch}{1.15}
\caption{Performance comparison on the FG-QA benchmark. 
Results are reported in percentage (\%). 
Alt. Acc. denotes Alternative Answer Set Accuracy, which measures the proportion of questions for which the predicted answer matches at least one of the alternative answers provided in the benchmark.}
\vspace{-3mm}
\label{tab:fg_qa_results}
\resizebox{\textwidth}{!}{
\begin{tabular}{c l c cccc cc cc c}
\toprule
\multirow{2}{*}{\textbf{\#}}
& \multirow{2}{*}{\textbf{Method}} 
& \multirow{2}{*}{\textbf{Overall Acc.}}
& \multicolumn{4}{c}{\textbf{queryRel}}
& \multicolumn{2}{c}{\textbf{existRel}}
& \multicolumn{2}{c}{\textbf{verifyRel}}
& \multicolumn{1}{c}{\textbf{chooseRel}} \\
\cmidrule(lr){4-7}
\cmidrule(lr){8-9}
\cmidrule(lr){10-11}
\cmidrule(lr){12-12}
&
&
&
Acc. & Alt. Acc. & WUPS@0.9 & WUPS@0.0
& Acc. & Macro-F1
& Acc. & Macro-F1
& Acc. \\
\midrule
\multicolumn{12}{l}{\textit{Multimodal Large Language Model}} \\
0 & Llama-3.2-11B-Vision-Instruct~\cite{grattafiori2024llama}
& 47.39
& 26.80
& 31.77
& 39.47
& 73.37
& 79.64
& 80.66
& 77.82
& 79.56
& 70.58 \\

1 & ALLaVA-3B~\cite{chen2024allava}
& 44.10
& 22.86
& 27.30
& 35.40
& 71.14
& 70.87
& 69.87
& 74.34
& 74.11
& 77.65 \\

2 & LLaVA-1.5-7B~\cite{liu2024improved}
& 48.49
& 26.05
& 30.80
& 38.59
& 73.08
& 79.62
& 79.60
& 78.90
& 78.86
& 74.11 \\

3 & Qwen2-VL-7B-Instruct~\cite{wang2024qwen2}
& 47.40
& 25.98
& 30.44
& 37.69
& 72.50
& 77.27
& 76.82
& 78.06
& 77.30
& 74.63 \\

4 & InternVL3.5-8B~\cite{wang2025internvl3}
& 44.24
& 18.94
& 29.10
& 36.24
& 71.13
& 81.38
& \underline{81.77}
& \underline{79.82}
& \underline{80.47}
& 66.14 \\

\midrule
\multicolumn{12}{l}{\textit{Specialized Methods for Fine-grained Visual Question Answering}} \\
5 & PinPoint (Qwen)~\cite{kwon2026focus}
& \underline{50.27}
& \underline{27.84}
& \underline{32.81}
& \underline{40.13}
& \underline{73.73}
& \underline{81.78}
& 81.68
& 79.76
& 79.76
& \underline{75.91} \\

6 & PinPoint (LLaVA)~\cite{kwon2026focus}
& 47.47
& 24.23
& 29.25
& 36.83
& 72.35
& 79.29
& 79.19
& 78.59
& 78.58
& 74.41 \\

\rowcolor{gray!20}
7 & SEAL~\citep{wu2024v}
& 29.24
& 12.75
& 15.72
& 23.05
& 63.04
& 46.05
& 37.44
& 48.95
& 38.41
& 60.90 \\

\rowcolor{gray!20}
8 & ViCrop~\cite{khayatkhoei2025mllms}
& 35.98
& 18.48
& 22.60
& 29.85
& 67.76
& 56.63
& 66.52
& 55.40
& 67.06
& 67.61 \\

\rowcolor{gray!20}
9 & Zoom Eye~\cite{shen2025zoomeye}
& 47.08
& 24.53
& 29.79
& 34.53
& 60.15
& 78.49
& 78.43
& 77.59
& 77.56
& 72.75 \\

\midrule
10 & Ours
& \textbf{59.59}
& \textbf{38.71}
& \textbf{42.61}
& \textbf{52.13}
& \textbf{85.39}
& \textbf{86.28}
& \textbf{89.73}
& \textbf{88.47}
& \textbf{88.56}
& \textbf{85.92} \\

\bottomrule
\end{tabular}
}
\vspace{-3mm}
\end{table*}

\begin{table*}[t]
\centering
\scriptsize
\setlength{\tabcolsep}{3.7pt}
\renewcommand{\arraystretch}{1.12}
\caption{Results on the FG-Caption.
Methods evaluated without fine-tuning, including training-free methods and some pretrained models used directly for inference, are highlighted in \graytext{gray}.
Best results are highlighted in \textbf{bold}, while second-best results are \underline{underlined}.
}
\vspace{-3mm}
\label{tab:fg_caption_results}
\begin{tabularx}{\textwidth}{c l c|*{8}{C}}
\toprule
 & \textbf{Method} & \textbf{LLM} 
 & \textbf{BLEU-1} & \textbf{BLEU-2} & \textbf{BLEU-3} & \textbf{BLEU-4}
 & \textbf{CIDEr} & \textbf{ROUGE-L} & \textbf{METEOR} & \textbf{CLIP-Score} \\
\midrule

\multicolumn{11}{l}{\textit{Open-source Multimodal Large Language Models}} \\
1 & LLaVA-1.5-7B~\cite{liu2024improved} & 7B
& 0.7294
& 0.5493
& 0.3938
& 0.2774
& 0.5381
& 0.4114
& 0.4164
& 0.6890 \\

2 & ALLaVA-3B~\cite{chen2024allava} & 3B
& 0.7009
& 0.5199
& 0.3643
& 0.2526
& 0.5079
& 0.4175
& 0.4085
& 0.6786 \\

3 & Llama-3.2-11B-Vision-Instruct~\cite{grattafiori2024llama}  & 11B
& 0.7476
& 0.5837
& 0.4160
& 0.3108
& 0.5554
& 0.4641
& 0.4688
& 0.6857 \\

4 & InternVL3.5-8B~\cite{wang2025internvl3} & 8B
& 0.7334
& 0.5790
& 0.4080
& 0.3047
& 0.5203
& 0.4476
& \underline{0.4742}
& 0.6967 \\

5 & Qwen2-VL-7B-Instruct~\cite{wang2024qwen2} & 7B
& 0.7380
& 0.5611
& 0.4117
& \underline{0.3263}
& \underline{0.6041}
& 0.4294
& 0.4608
& 0.6993 \\

\midrule
\multicolumn{11}{l}{\textit{Specialized Methods for Fine-grained Image Captioning}} \\
\rowcolor{gray!20}
6 & IT~\cite{pi2024image} & 7B
& 0.3968
& 0.2509
& 0.1439
& 0.0808
& 0.1329
& 0.2568
& 0.3259
& 0.6416 \\

\rowcolor{gray!20}
7 & Patch Matters~\cite{peng2025matters}  & 7B
& 0.5061
& 0.3406
& 0.2141
& 0.1328
& 0.1757
& 0.3117
& 0.3917
& \textbf{0.7057} \\

\rowcolor{gray!20}
8 & Synthesized~\cite{dong2024benchmarking} & 7B
& 0.4517 & 0.2830 & 0.1675 & 0.0984
& 0.1269 & 0.2623 & 0.3459 & 0.6783 \\

9 & ShareGPT4V-7B~\cite{chen2024sharegpt4v} & 7B
& \underline{0.7582} & \underline{0.5851} & \underline{0.4329} & 0.3162
& 0.5972 & \textbf{0.5337} & 0.4373 & 0.6921 \\

10 & SC-Captioner~\cite{zhang2025sc} & 7B
& 0.7319 & 0.5762 & 0.4008 & 0.2995
& 0.5874 & \underline{0.4836} & 0.4596 & 0.6916 \\

\midrule
11 & FiGRA \textit{(ours)}   & 7B  
& \textbf{0.7613} 
& \textbf{0.5928}
& \textbf{0.4477} 
& \textbf{0.3293}
& \textbf{0.6265}
& 0.4788
& \textbf{0.4791}
& \underline{0.7011} \\
\bottomrule
\end{tabularx}
\vspace{-3mm}
\end{table*}

\subsection{FiGRA ablation study}
\label{sec:figra_ablation}

To answer \textbf{Q2}, we evaluate four key components of FiGRA. Results are reported in \Cref{tab:ablation}.

\emph{\textbf{(1) w/o Shortest-Path Prediction.}}
We remove shortest-path prediction from pre-training while retaining feature and topology reconstruction. Performance drops across all three benchmarks. 
This confirms that shortest-path supervision provides relational signals beyond feature and topology reconstruction.
\emph{\textbf{(2) w/o Pre-training.}}
We remove pre-training stage and train randomly initialized graph modules only during supervised fine-tuning. 
Performance drops across all three benchmarks. 
This confirms the importance of pre-training for learning effective entity and relation representations.
\emph{\textbf{(3) w/o Subgraph Retrieval.}}
We replace question-relevant subgraph retrieval with the complete MSG for FG-QA and FG-Caption. FG-QA accuracy drops from 59.59\% to 53.09\%. This highlights the importance of selecting relevant entities and relations for question answering.
\emph{\textbf{(4) w/o Graph Encoder.}}
We remove the Graph Encoder and directly apply graph-to-sequence transformation to the fused entity and relation features. Performance drops across all three benchmarks. This confirms the importance of graph message passing for incorporating relational context into entity representations.

\begin{table}[t]
\centering
\caption{Ablation study of FiGRA on the three tasks of FG-GraphBench. BLEU-4 is reported for FG-Description and FG-Caption, while overall accuracy (\%) is reported for FG-QA.}
\label{tab:ablation}
\small
\setlength{\tabcolsep}{3pt}
\renewcommand{\arraystretch}{1.15}
\resizebox{\columnwidth}{!}{
\begin{tabular}{lccc}
\toprule
Configuration & \shortstack{FG-\\Description\\BLEU-4} & \shortstack{FG-\\Caption\\BLEU-4} & \shortstack{FG-QA\\Overall Acc. (\%)} \\
\midrule
A. w/o Shortest-Path Prediction & 0.3594 & 0.3228 & 58.20 \\
B. w/o Pre-training & 0.3522 & 0.3217 & 57.64 \\
C. w/o Subgraph Retrieval & 0.3611 & 0.3002 & 53.09 \\
D. w/o Graph Encoder & 0.3509 & 0.3257 & 57.91 \\
\midrule
E. Full FiGRA & 0.3611 & 0.3293 & 59.59 \\
\bottomrule
\end{tabular}
}
\vspace{-3mm}
\end{table}

\begin{figure*}[t]
\centering
\includegraphics[width=\linewidth]{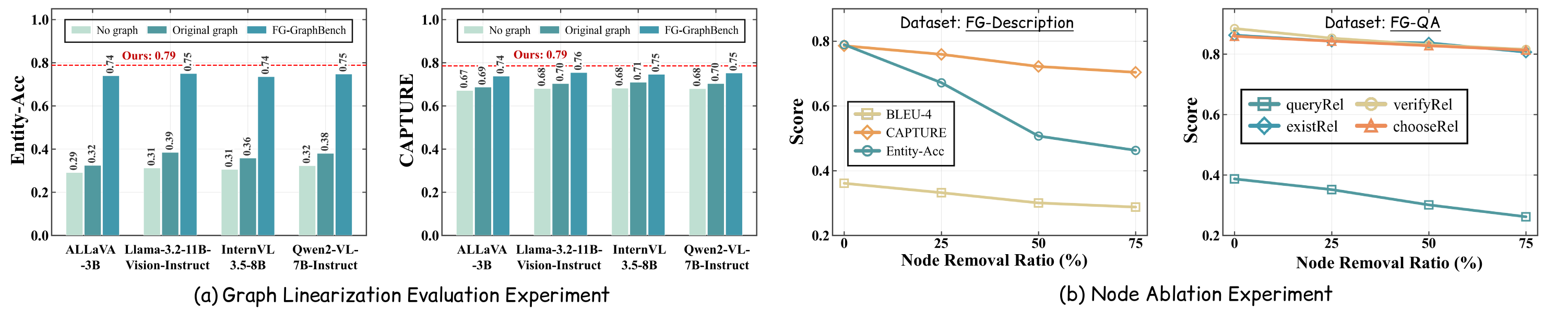}
\vspace{-5mm}
\caption{
Effectiveness of FG-GraphBench. (a) Graph linearization evaluation experiment on FG-Description. 
(b) Node ablation experiment on FG-Description (left) and FG-QA (right).
}
\label{fig:q3}
\vspace{-3mm}
\end{figure*}

\subsection{Effectiveness of FG-GraphBench}
\label{sec:fg_graphbench_utility}
To answer \textbf{Q3}, we conduct graph linearization evaluation and node ablation experiments to evaluate the utility of FG-GraphBench.

\textbf{Graph Linearization Evaluation Experiment.}
We evaluate four open-source MLLMs on FG-Description under three input settings:
(i) \emph{\textbf{No Graph}}, where the model receives only the image; (ii) \emph{\textbf{Original Graph}}, where the original scene graph from Action Genome is serialized into text and provided to MLLMs; and (iii) \emph{\textbf{FG-GraphBench}}, where the fine-grained MSG constructed by our data pipeline is serialized in the same way. 
As shown in~\cref{fig:q3}(a), FG-GraphBench consistently achieves the highest Entity-Acc and CAPTURE scores across all four MLLMs, with particularly substantial improvements in Entity-Acc. 
These results show that FG-GraphBench provides additional fine-grained entity and relation information beyond the original scene graphs, improving image recognition and description.

\textbf{Node Ablation Experiment.}
We further conduct node ablation experiments on FG-Description and FG-QA to assess the contribution of graph information.
For each sample, we obtain graphs with four node removal ratios (0\%, 25\%, 50\%, and 75\%) by removing the corresponding proportion of nodes from its complete fine-grained MSG.
When a node is removed, all edges connected to it are also removed. 
We evaluate FiGRA performance on FG-Description and FG-QA. 
As shown in~\cref{fig:q3}(b), increasing the node removal ratio consistently degrades performance on both tasks. 
These results confirm that our fine-grained MSGs provide useful visual information for tasks.

\subsection{Effect of graph information injection methods}
\label{sec: q4}
To address \textbf{Q4}, we compare three graph information injection methods by downstream performance and computational cost.

\begin{figure}[t]
\centering
\includegraphics[width=\linewidth]{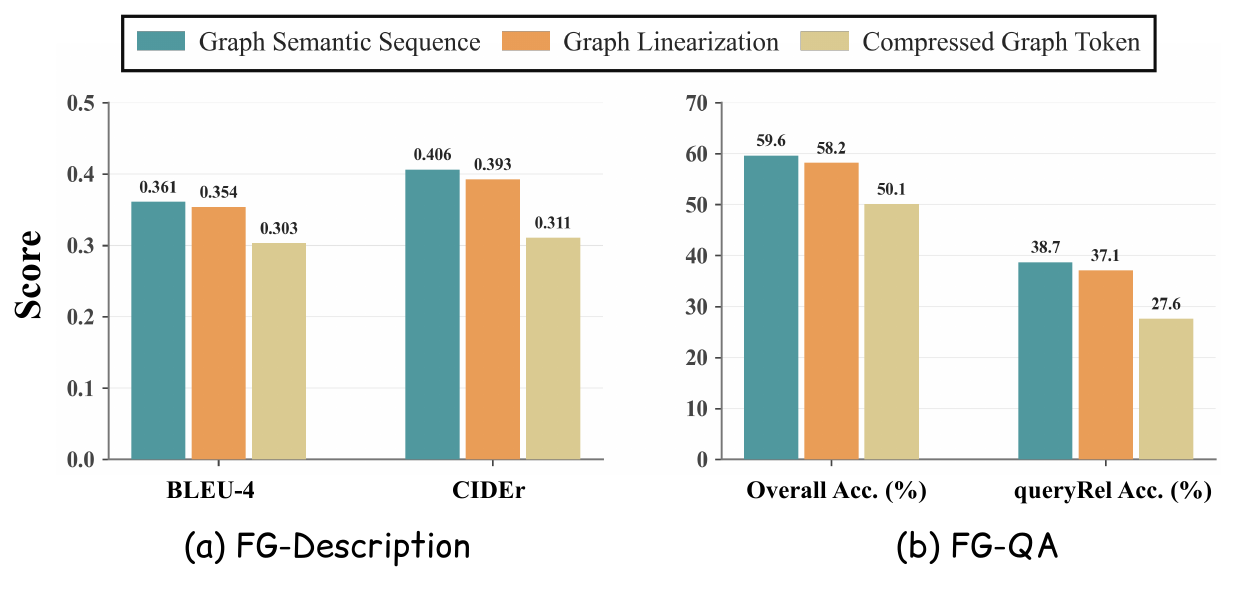}
\vspace{-3mm}
\caption{
Performance comparison of different graph information injection methods. 
}
\label{fig:q4}
\vspace{-4mm}
\end{figure}

\begin{table}[H]
    \centering
    \caption{
    Computational cost and module configurations of different graph information injection methods. \textbf{gr.}, \textbf{ge.}, and \textbf{sa.} denote graph retrieval, graph encoder, and semantic adapter, respectively. 
    }
    \label{tab:graph_injection_efficiency}
    \small
    \setlength{\tabcolsep}{4pt}
    \renewcommand{\arraystretch}{1.12}
    \resizebox{\columnwidth}{!}{%
    \begin{tabular}{lccc rrr}
        \toprule
        Method
        & gr. & ge. & sa.
        & \shortstack{Avg. input\\tokens $\downarrow$}
        & \shortstack{training\\hours/epoch $\downarrow$}
        & \shortstack{Inference\\hours $\downarrow$} \\
        \midrule
        Graph Linearization
        & $\checkmark$ & -- & --
        & 1063 & 4.09 & 5.48 \\
        Compressed Graph Token
        & $\checkmark$ & $\checkmark$ & $\checkmark$
        & \textbf{668} & \textbf{3.20} & \textbf{5.00} \\
        \rowcolor{gray!15}
        Graph Semantic Sequence (Ours)
        & $\checkmark$ & $\checkmark$ & $\checkmark$
        & 782 & 3.35 & 5.20 \\
        \bottomrule
    \end{tabular}%
    }
\end{table}

\textbf{Performance Comparison.}
We compare Graph Semantic Sequence (Ours), Graph Linearization, and Compressed Graph Token on FG-Description and FG-QA.
All three methods use the same graph information for each benchmark.
Graph Semantic Sequence (Ours) uses the complete graph semantic sequence as described in~\cref{sec:graph_token_integration}.
Graph Linearization converts the graph into text without the Graph Encoder and Semantic Adapter.
Compressed Graph Token averages all node and edge embeddings produced by the Graph Encoder into a single vector through mean pooling, which is then projected by the Semantic Adapter and injected into the MLLM as a single graph token.

\textbf{Computational Cost.}
We report average input tokens, training time per epoch, and inference time. 
All methods use the same graph retrieval strategy. 

As shown in~\Cref{fig:q4} and~\Cref{tab:graph_injection_efficiency}, Graph Linearization causes only marginal performance drops but incurs the highest computational costs.
This suggests that MLLMs can effectively interpret graph information in text form, but this representation achieves slightly lower performance while requiring more computational resources. 
In contrast, Compressed Graph Token achieves the lowest computational costs but suffers substantial performance degradation, particularly on FG-QA, suggesting that mean pooling discards fine-grained entity and relation information. 
These results reveal a trade-off between computational efficiency and fine-grained visual understanding.
Our method achieves a better balance between the two.

\section{Conclusion}
To address the limited coverage of fine-grained entities and their relationships in existing scene graph benchmarks, in this paper, we introduced FG-GraphBench, a benchmark for fine-grained visual understanding across three complementary tasks with fine-grained MSGs. 
We developed a two-stage data construction pipeline to construct the fine-grained MSGs in FG-GraphBench by discovering fine-grained visual entities and establishing semantic and dependency relations among them.
To address the challenge of effectively incorporating fine-grained graph information into MLLMs, we proposed FiGRA, which integrates multimodal graph representations into MLLMs through continuous graph tokens and a two-stage training strategy. 
Experiments show that FiGRA outperforms existing methods across all three fine-grained visual understanding tasks. 
Further analyses highlight the importance of explicitly modeling fine-grained entities and their relationships in MLLMs for fine-grained visual understanding.


\section*{LLM Usage Statement}
In our data construction pipeline, we used Qwen3.7-Plus for data annotation.
In preparing this paper, we used ChatGPT (OpenAI) to polish drafts written by ourselves, including shortening or expanding sentences and correcting grammar, and to help identify relevant literature for the Related Work section. All suggested revisions were manually reviewed and selectively incorporated based on the authors’ judgment. ChatGPT was not used for experimental design and model architecture development.

{
    \small
    \bibliographystyle{ieeenat_fullname}
    \bibliography{FG-GraphBench}
}

\clearpage
\onecolumn
\setcounter{page}{1}

\begin{center}
    {\Large\bfseries \thetitle\par}
    \vspace{0.5em}
    {\large Supplementary Material\par}
    \vspace{1em}
\end{center}
\appendix

\section{Dataset Information}
\label{app:dataset-information}

\subsection{Dataset Scale}
\label{app:dataset-scale}

\textbf{Action Genome (AG)~\cite{ji2020action}.}
Action Genome is a large-scale video dataset built on Charades to represent human-object activities as spatio-temporal scene graphs. Its annotations localize the acting person and the objects involved in each action with bounding boxes, and describe their interactions through attention, spatial, and contact relations. The annotation protocol uniformly samples five frames from each action interval, providing frame-level graphs that capture how interactions evolve over time. 
For our FG-Description, we select one representative frame per video, corresponding to the frame with the largest number of visible annotated objects.
The scene graph annotations associated with these frames serve as the original scene graphs.

\textbf{MS COCO Captions~\cite{chen2015microsoft}.}
MS COCO Captions is an image-captioning benchmark built on the MS COCO image collection, which was designed to include multiple objects in natural scene contexts. 
The caption dataset itself does not contain scene-graph annotations. 
Therefore, for FG-Caption, we associate each COCO image with its overlapping GQA image and reuse the corresponding GQA scene graph, including objects, bounding boxes, and relations.

\textbf{GQA~\cite{hudson2019gqa}.}
GQA is a compositional visual question answering dataset built upon Visual Genome scene graphs covering approximately 113K real-world images~\cite{krishna2017visual}. 
Each graph represents visible objects as nodes, with bounding boxes and typically one to three attributes, and connects them through semantic relations describing actions, spatial configurations, and comparisons. 
For FG-QA, we exclude GQA images that overlap with MS COCO Captions and retain only the scene graphs associated with the remaining images as our original scene graphs.

We quantify the changes in graph size introduced by our data
construction pipeline. 
For each dataset, we compare the original scene graphs
with our fine-grained graphs.
\Cref{tab:dataset-node-statistics} reports the number of nodes before
and after our data construction pipeline.
\Cref{tab:dataset-edge-statistics} reports the corresponding edge statistics.

\begin{table}[H]
\centering
\small
\setlength{\tabcolsep}{4pt}
\renewcommand{\arraystretch}{1.12}
\begin{tabular}{lrrrr}
\toprule
Dataset & Original & Ours & Relative change & Avg. nodes (original/ours) \\
\midrule
\textsc{FG-Description} & 35,942 & 155,373 & +332.29\% & 3.75 / 16.19 \\
\textsc{FG-Caption} & 228,671 & 491,677 & +115.02\% & 16.30 / 35.05 \\
\textsc{FG-QA} & 740,088 & 1,244,016 & +68.09\% & 17.06 / 28.67 \\
\bottomrule
\end{tabular}%
\caption{Comparison of node counts between the original scene graphs and the graphs constructed by our data construction pipeline. 
Relative change is calculated with respect to the original count. Average node counts are reported per graph.}
\label{tab:dataset-node-statistics}
\end{table}

\begin{table}[H]
\centering
\small
\setlength{\tabcolsep}{4pt}
\renewcommand{\arraystretch}{1.12}
\begin{tabular}{lrrrr}
\toprule
Dataset & Original & Ours & Relative change & Avg. edges (original/ours) \\
\midrule
\textsc{FG-Description} & 84,410 & 336,586 & +298.75\% & 8.80 / 35.08 \\
\textsc{FG-Caption} & 685,803 & 1,421,056 & +107.21\% & 48.90 / 101.32 \\
\textsc{FG-QA} & 2,320,725 & 3,523,486 & +51.83\% & 53.49 / 81.22 \\
\bottomrule
\end{tabular}%
\caption{Comparison of edge counts between the original scene graphs and the graphs constructed by our data construction pipeline. Relative change is calculated with respect to the original count. Average edge counts are reported per graph.}
\label{tab:dataset-edge-statistics}
\end{table}

\subsection{Node and Edge Composition}
\label{app:graph-composition}

\begin{figure}[t]
\centering
\includegraphics[width=\linewidth]{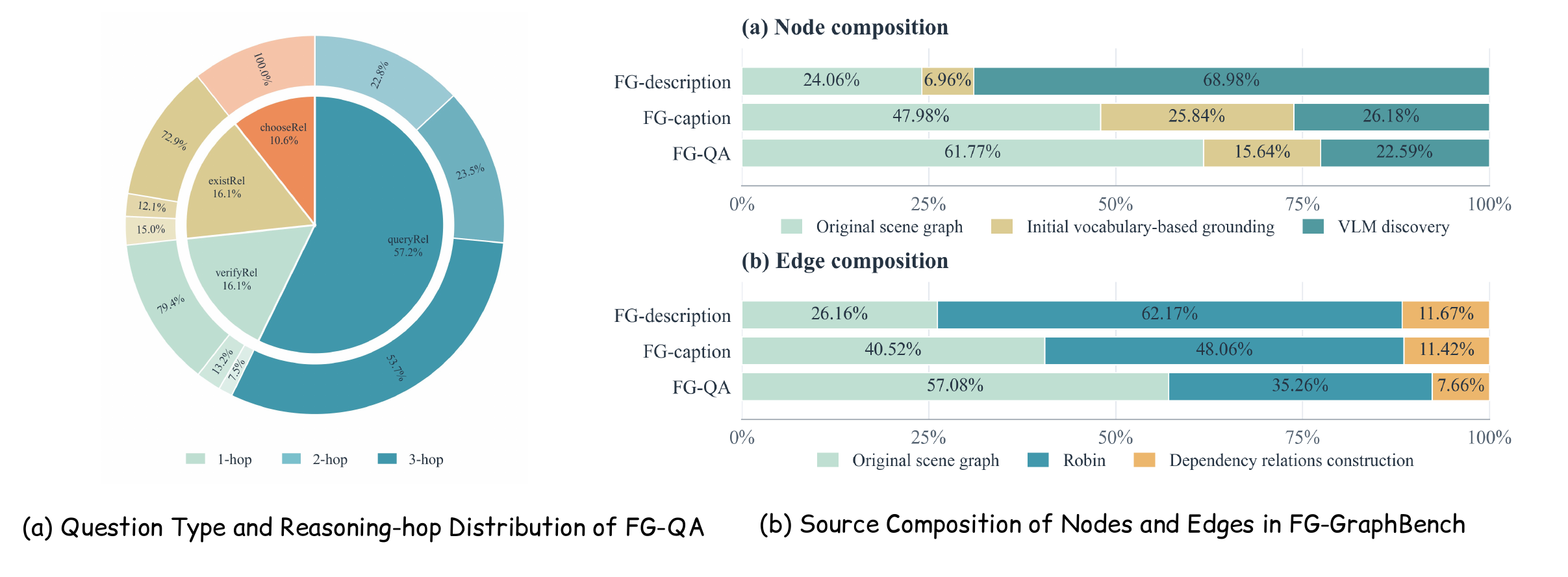}
\caption{
Question type and reasoning-hop distributions in \textsc{FG-QA} (a),
and the composition of our fine-grained scene graph nodes and edges (b).
The inner ring in (a) shows the proportions of question types, the outer ring shows the hop distribution within each type.
}
\label{fig:graph-composition-and-hops}
\end{figure}

\Cref{fig:graph-composition-and-hops}(b) shows the composition of nodes and edges from each source across the three datasets.

\textbf{Node composition.}
We categorize nodes into three groups according to their sources: Original scene graph, Initial vocabulary-based grounding, and VLM discovery. 
The first group consists of entities retained from the original scene graphs. 
The second comprises additional entities obtained through initial vocabulary-based grounding.
The third comprises entities discovered through the two-round VLM process and subsequently grounded. 


\textbf{Edge composition.}
We classify the edges into three categories: original scene graph relations, relations generated by ROBIN, and dependency relations constructed over the Dependency Candidate Tree.

\subsection{Question Types and Reasoning Hops of FG-QA}
\label{app:fgqa-reasoning-complexity}

\textsc{FG-QA} contains four question types: queryRel, verifyRel, existRel, and chooseRel.
Table~\ref{tab:fgqa-hop-distribution} reports the number of questions of each type and the distribution of their reasoning hops.

Reasoning hops refer to the number of relation steps traversed in a reasoning chain. 
For example, a 2-hop question is \textit{person} $\xrightarrow{\text{holding}}$ \textit{cup} $\xrightarrow{\text{on}}$ \textit{table}, corresponding to ``Where is the cup held by the person?'' 
The hop count characterizes the depth of relational reasoning required to answer a question.

\begin{table}[t]
\centering
\small
\setlength{\tabcolsep}{4pt}
\renewcommand{\arraystretch}{1.12}
\begin{tabular}{lrrrr}
\toprule
Question type & Questions & 1-hop & 2-hop & 3-hop \\
\midrule
queryRel & 24,824 & 5,665 {\scriptsize (22.82\%)} & 5,837 {\scriptsize (23.51\%)} & 13,322 {\scriptsize (53.67\%)} \\
verifyRel & 6,977 & 522 {\scriptsize (7.48\%)} & 918 {\scriptsize (13.16\%)} & 5,537 {\scriptsize (79.36\%)} \\
existRel & 6,974 & 1,049 {\scriptsize (15.04\%)} & 842 {\scriptsize (12.07\%)} & 5,083 {\scriptsize (72.89\%)} \\
chooseRel & 4,609 & 4,609 {\scriptsize (100.00\%)} & 0 {\scriptsize (0.00\%)} & 0 {\scriptsize (0.00\%)} \\
\midrule
All & 43,384 & 11,845 {\scriptsize (27.30\%)} & 7,597 {\scriptsize (17.51\%)} & 23,942 {\scriptsize (55.19\%)} \\
\bottomrule
\end{tabular}
\caption{Reasoning-hop distribution of \textsc{FG-QA}. Percentages are computed within each question type.}
\label{tab:fgqa-hop-distribution}
\end{table}

\subsection{Answer Distribution of FG-QA}
\label{app:fgqa-answer-distribution}

We analyze the answer distributions of the verifyRel and existRel questions and the open-ended queryRel questions separately.

\textbf{Yes/no answers.}
Table~\ref{tab:fgqa-yes-no} summarizes the answer distributions for verifyRel and existRel.
Both question types exhibit nearly balanced yes/no distributions.
For verifyRel, yes and no answers account for 49.05\% and
50.95\% of the questions, respectively. The corresponding proportions for existRel are 49.07\% and 50.93\%.

\begin{table}[t]
\centering
\small
\setlength{\tabcolsep}{8pt}
\renewcommand{\arraystretch}{1.12}
\begin{tabular}{lrrr}
\toprule
Question type & Questions & Yes & No \\
\midrule
verifyRel & 6,977 & 3,422 {\scriptsize (49.05\%)} & 3,555 {\scriptsize (50.95\%)} \\
existRel & 6,974 & 3,422 {\scriptsize (49.07\%)} & 3,552 {\scriptsize (50.93\%)} \\
\bottomrule
\end{tabular}
\caption{Yes/no answer distributions in \textsc{FG-QA}. Percentages are computed within each question type.}
\label{tab:fgqa-yes-no}
\end{table}

\begin{figure}[t]
\centering
\includegraphics[width=\linewidth]{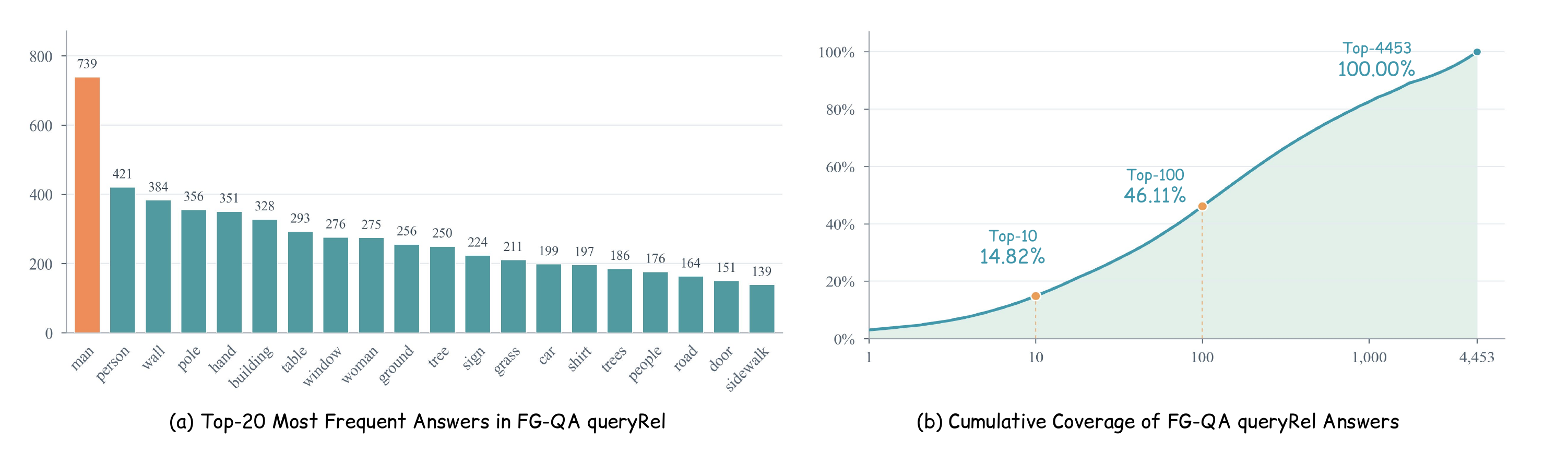}
\caption{
Answer distribution of queryRel in FG-QA.
(a) shows the frequencies of the 20 most common answers.
(b) shows the cumulative coverage of FG-QA queryRel answers.
}
\label{fig:fgqa-answer-distribution}
\end{figure}

\textbf{Open-ended answers.}
For queryRel, the answer field contains 4,453 distinct answer strings across 24,824 questions.
Table~\ref{tab:fgqa-answer-coverage} reports the proportion of questions covered by the most frequent answer categories. 
The Top-$K$ coverage is defined as the number of questions assigned to the $K$ most frequent answer categories divided by the total number of queryRel questions.
\Cref{fig:fgqa-answer-distribution} visualizes the frequencies of the 20 most common answers and the cumulative coverage of FG-QA queryRel answers.

\begin{table}[t]
\centering
\small
\setlength{\tabcolsep}{8pt}
\renewcommand{\arraystretch}{1.12}
\begin{tabular}{rrr}
\toprule
Top-$K$ categories & Questions covered & Coverage (\%) \\
\midrule
Top-1   & 739    & 2.98 \\
Top-5   & 2,251  & 9.07 \\
Top-10  & 3,679  & 14.82 \\
Top-20  & 5,576  & 22.46 \\
Top-50  & 8,497  & 34.23 \\
Top-100 & 11,446 & 46.11 \\
\bottomrule
\end{tabular}
\caption{Cumulative coverage of the most frequent answer categories in queryRel.}
\label{tab:fgqa-answer-coverage}
\end{table}

\subsection{Dataset Splits.}
\label{app:dataset-splits-protocol}

We split each dataset into training, validation, and test sets at a ratio of 6:2:2. The training set is used for model training, the test set for model selection, and the validation set for final evaluation. 
\Cref{tab:dataset-splits} reports the number of samples in each split.

\begin{table}[t]
\centering
\small
\setlength{\tabcolsep}{7pt}
\renewcommand{\arraystretch}{1.12}
\begin{tabular}{lrrr}
\toprule
Dataset & Train & Validation & Test \\
\midrule
\textsc{FG-Description} & 5,757 & 1,918 & 1,920 \\
\textsc{FG-Caption} & 8,415 & 2,805 & 2,806 \\
\textsc{FG-QA} & 26,030 & 8,676 & 8,678 \\
\bottomrule
\end{tabular}
\caption{Dataset split statistics.}
\label{tab:dataset-splits}
\end{table}

\subsection{Data Quality Check}
\label{sec:human_quality_audit}

To evaluate the reliability of the automatically constructed fine-grained scene graphs, we conduct a human inspection of three datasets: FG-QA (GQA), FG-Description (Action Genome), and FG-Caption (COCO). The data quality check focuses on the correctness of newly introduced entities and relations, particularly those involving fine-grained objects.

\paragraph{Data Quality Check Protocol.}
For each dataset, we divide the sampled images into two groups according to the number of newly added nodes relative to the original scene graph. Specifically, a sample is categorized as \textit{simple} if the number of newly added nodes is smaller than the number of nodes in its original scene graph, indicating relatively limited graph expansion. Otherwise, a sample is categorized as \textit{high-risk} if the number of newly added nodes exceeds the number of original nodes, indicating more substantial graph expansion and potentially greater annotation uncertainty.

Human annotators inspect the newly introduced nodes and edges against the original images. For each new node, they verify whether its semantic name correctly matches the corresponding visual region and mask. For each new edge, they check whether its relation type and direction are consistent with the visual evidence in the image.

We report four metrics to assess annotation quality:
\begin{itemize}
    \item \textbf{Entity Accuracy ($Acc_{\mathrm{entity}}$):} The proportion of newly added nodes whose names correctly match their corresponding masks.
    \item \textbf{Relation Accuracy ($Acc_{\mathrm{relation}}$):} The proportion of newly added edges with both correct relation labels and directions.
    \item \textbf{Fine-grained Entity Accuracy ($Acc_{\mathrm{fg\text{-}entity}}$):} The proportion of correctly annotated fine-grained entities, defined as newly added nodes whose mask areas occupy less than 1\% of the entire image area.
    \item \textbf{Fine-grained Relation Accuracy ($Acc_{\mathrm{fg\text{-}relation}}$):} The proportion of correctly annotated fine-grained relations, defined as newly added edges for which at least one endpoint corresponds to a fine-grained entity.
\end{itemize}

\paragraph{Overall Data Quality.}
Table~\ref{tab:human_audit_overall} presents the data quality check results for all newly introduced entities and relations. Entity accuracy remains consistently high across the three datasets. 
In comparison, relation accuracy ranges from 83.4\% to 86.0\%, indicating that relation construction is generally more challenging than entity grounding.

\begin{table*}[t]
\centering
\small
\setlength{\tabcolsep}{7pt}
\begin{tabular}{llccc}
\toprule
Dataset & Group & \#Samples
& $Acc_{\mathrm{entity}}$ & $Acc_{\mathrm{relation}}$ \\
\midrule
\multirow{3}{*}{FG-QA (GQA)}
& Simple & 50 & 90.1\% & 87.7\% \\
& High-risk & 50 & 89.3\% & 84.2\% \\
& Overall & 100 & 89.6\% & 85.6\% \\
\midrule
\multirow{3}{*}{FG-Description (AG)}
& Simple & 24 & 87.4\% & 88.9\% \\
& High-risk & 50 & 88.1\% & 82.0\% \\
& Overall & 74 & 88.0\% & 83.4\% \\
\midrule
\multirow{3}{*}{FG-Caption (COCO)}
& Simple & 50 & 90.8\% & 88.3\% \\
& High-risk & 50 & 90.2\% & 83.9\% \\
& Overall & 100 & 90.4\% & 86.0\% \\
\bottomrule
\end{tabular}
\caption{Data quality check results for newly introduced entities and relations in FG-GraphBench. Overall accuracies are computed by aggregating the inspected entities or relations across the two groups.}
\label{tab:human_audit_overall}
\end{table*}

\begin{table*}[t]
\centering
\small
\setlength{\tabcolsep}{7pt}
\begin{tabular}{llccc}
\toprule
Dataset & Group & \#Samples
& $Acc_{\mathrm{fg\text{-}entity}}$
& $Acc_{\mathrm{fg\text{-}relation}}$ \\
\midrule
\multirow{3}{*}{FG-QA (GQA)}
& Simple & 50 & 85.4\% & 82.6\% \\
& High-risk & 50 & 86.3\% & 81.7\% \\
& Overall & 100 & 86.0\% & 82.0\% \\
\midrule
\multirow{3}{*}{FG-Description (AG)}
& Simple & 24 & 84.9\% & 83.9\% \\
& High-risk & 50 & 85.1\% & 78.2\% \\
& Overall & 74 & 85.1\% & 79.8\% \\
\midrule
\multirow{3}{*}{FG-Caption (COCO)}
& Simple & 50 & 86.6\% & 84.4\% \\
& High-risk & 50 & 85.1\% & 80.2\% \\
& Overall & 100 & 85.6\% & 82.2\% \\
\bottomrule
\end{tabular}
\caption{Data quality check results for fine-grained entities and relations. Fine-grained entities are defined by a mask area smaller than 1\% of the image area, and fine-grained relations are edges involving at least one such entity.}
\label{tab:human_audit_finegrained}
\end{table*}

\paragraph{Fine-grained Data Quality.}
Since fine-grained entities are typically smaller and more difficult to identify and localize, we further evaluate the accuracy of entities with mask areas smaller than 1\% of the image and the relations involving these entities. Table~\ref{tab:human_audit_finegrained} reports the corresponding accuracy.

\paragraph{Analysis.}
As shown in Table~\ref{tab:human_audit_overall}, the newly introduced entities and relations achieve consistently high accuracy across all three datasets, with overall entity accuracy ranging from 88.0\% to 90.4\% and relation accuracy from 83.4\% to 86.0\%, demonstrating the reliability of our graph construction pipeline. 
However, as shown in Table~\ref{tab:human_audit_finegrained}, the accuracy decreases for fine-grained entities and their associated relations, indicating that small objects and the relations involving them remain more challenging to identify and annotate accurately.

Overall, the data quality check confirms that FG-GraphBench provides reliable fine-grained entity and relation annotations across three different source datasets. 
Meanwhile, the analysis identifies relations involving fine-grained entities as a more challenging component.

\subsection{Question Construction}
\label{Question Construction}

\textbf{FG-QA question construction.}
Inspired by the relational question templates in GQA~\citep{hudson2019gqa}, we generate four types of questions from graph relations: queryRel asks for an entity connected to another entity by a specified relation, verifyRel verifies whether a specified relation holds, existRel checks whether an entity satisfying given relational constraints exists, and chooseRel selects the correct relation from two candidate spatial relations. The first three types can involve multi-hop entity references, whereas chooseRel focuses on direct relations between two entities.

We construct entity references from graph paths and filter candidates based on path validity, reasoning depth, and the relative mask areas of the involved entities. 
Negative examples for verifyRel and existRel are generated by modifying entities or relations to produce statements unsupported by the graph. 
For each image, we first sample a question type using weights of 0.60, 0.15, 0.15, and 0.10 for queryRel, verifyRel, existRel, and chooseRel, respectively. 
We then retain up to five candidates of the selected type, prioritizing longer reasoning paths and higher mask area scores.

Finally, we use Qwen3.7-Plus to revise questions that are grammatically incorrect or unnatural, while preserving their original meaning.
For questions that require an entity or relation as the answer, the VLM also proposes alternative answers with equivalent meanings. The ground-truth answer remains determined by the verified graph structure rather than generated by the VLM.

\textbf{FG-Description question construction.}
For FG-Description, we formulate question construction as a detailed image description task. 
We provide the Qwen3.7-Plus with the image. 
We generate three complementary descriptions for each sample using different instructions. 
The first produces a faithful overall account, the second emphasizes visible fine-grained details, and the third focuses on spatial layout, background, and scene context.
All variants are required to form a single fluent English paragraph, balance the main subjects with informative details, and describe only observable content.

\textbf{FG-Caption question construction.}
FG-Caption is constructed by refining the five human-written COCO captions associated with each image. 
The original captions and the image are provided jointly to the VLM (Specifically, we use Qwen3.7-Plus). 
Each source caption is rewritten independently while preserving its main meaning. 
The VLM is encouraged to add visually supported information about objects, attributes, materials, clothing, body parts, actions, background elements, and spatial relations.

\section{Detailed Implementation of Stage 1}
\label{app:grounded-entity-construction}

\subsection{Initial Vocabulary-based Grounding}
\label{app:initial-vocabulary-grounding}

The main paper describes how the source vocabulary initializes grounding. Here we give the implementation details.
For every source entity, its annotated bounding box is clipped to the image boundary and used as a box prompt for SAM.
We use the ViT-H variant of SAM and request a single mask for each box.

We use GroundingDINO with the Swin-T OGC checkpoint~\citep{liu2024grounding} to localize each distinct source entity name in an image, with box and text thresholds of $0.30$ and $0.25$, respectively. The resulting boxes are segmented with SAM ViT-H, and the masks are stored separately from the source masks to preserve the original annotations.

\subsection{Two Round VLM Fine-grained Entity Discovery}
\label{app:two-round-vlm-discovery}

We use Qwen3.7-plus for both rounds via the DashScope API, with the maximum output length set to $10{,}000$ tokens. The model receives the original image together with a simplified source scene graph.

In Round~1, the model is required to select at least five source entities as anchors and to return at least three previously unlisted fine-grained components for each selected anchor. Each component includes a short name and a visual description. See Table~\ref{tab:round1-prompt} for our Round~1 prompt.

\begin{table*}[!t]
\centering
\begin{promptbox}[Round 1 prompt]
I will provide an ``Image'' and a ``Scene Graph'' containing the major entities already present in the image. The Scene Graph is guaranteed to be correct.\\[2pt]
\relax[Image]\\
- The image is a real-world scene with no fictional content.\\
- All entities come from real life.\\
\relax[Scene Graph]\\
A scene graph is a graph structure used to describe the visual scene in this image. The scene graph structure is:\\
\textcolor{red}{\{graph\_format\_text\}}\\
\normalfont\small
\textbf{Objective:}Your task is to expand the granularity of the current scene graph by identifying micro-scale objects and fine-grained sub-parts.\\
\textbf{1. Anchor Selection:}
Observe the image and cross-reference it with the [Scene Graph]. Select at least 5 primary entities that are visually prominent, occupy significant regions, or possess high informational density (e.g., a `person' in the center, a `vehicle' in the foreground, or a `complex machine'). These will serve as anchors for detailed decomposition.\\
\textbf{Important:} every anchor must be an existing entity in the provided [Scene Graph], and you must output its exact node id as `anchor\_entity\_id'.\\
\textbf{2. Micro-Object Discovery:} For each selected entity, you must identify at least 3 specific fine-grained components, small attachments, or micro-objects that are clearly visible but NOT yet listed in the [Scene Graph] (e.g., if the entity is a `person', look for `glasses', `watch', 'buttons', or 'sneakers').\\
\textbf{3. Evidence Extraction:} For every identified micro-object, provide a concise visual description of its appearance.\\
\textbf{4. Naming Rule:} Keep each `name' short and use common everyday words. ``Prefer a noun phrase with one visible attribute when possible (not mandatory), from: `color / material / pattern / shape / size / state`. If you add an attribute, use only one attribute word."\\
\textbf{Example:}: use `tall building` or `gray building' instead of `tall gray concrete building'.\\
Output pattern:\\
{``results": \\
\relax[``anchor\_entity\_id": 12,``anchor\_entity":``person",\\
``micro\_objects": [``name": "white button", ``description": ``small round white buttons on the front"},{``name": ``black watch", ``description": ``black leather strap watch on the left wrist"},{``name": ``folded collar", ``description": ``folded fabric edge at the neck of the shirt"}]
]\\
\textbf{Your output must be JSON only (no markdown).}\\
\textbf{Data:}Allowed anchor entities (must use ids from this list):\textcolor{red}{\{anchor\_candidates\_text\}},\\
scene graph:\textcolor{red}{\{graph\_text\}}.
\end{promptbox}
\caption{Prompt used for first-round fine-grained entity discovery.}
\label{tab:round1-prompt}
\end{table*}

Round~2 receives the image and source scene graph, together with the fine-grained entities identified in Round~1. This additional context helps avoid duplicate proposals and directs the model toward entities not yet covered. The implementation requests at least five additional objects, with particular attention to small entities. See Table~\ref{tab:round2-prompt} for our Round~2 prompt.

\begin{table*}[!t]
\centering
\begin{promptbox}[Round 2 prompt: global discovery]
I will provide an ``Image", a Sub-part List and a Scene Graph containing the major entities already present in the image
The ``Scene Graph" is guaranteed to be correct.\\
\relax[Image]\\
- The image is a real-world scene with no fictional content.\\
- All entities come from real life.\\
\relax[Scene Graph]\\
A scene graph is a graph structure used to describe the visual scene in this image. The scene graph structure is:\\
\textcolor{red}{\{graph\_format\_text\}}\\
\relax[Sub-part List]\\
The sub-part list contains fine-grained components that were already identified and should be treated as known items.\\
The sub-part list structure is:\\
\textcolor{red}{\{sub\_part\_format\_text\}}\\
\normalfont\small
\textbf{Objective:} Your task is to identify any additional independent entities in the image that have been completely missed so far. These objects are not entities in the [Scene Graph], nor are they entities in the provided [Sub-part List], but are clearly visible and contribute to the complete understanding of the scene (especially small, subtle, partially occluded, distant, or easily overlooked items such as tiny foreground/background objects and fine-grained parts).\\
\textbf{1. Global Scanning:} Scan the entire image for any visual elements (entities or parts) that are missing from both lists provided above.Pay special attention to small, subtle, distant, partially occluded, edge-region, and easily overlooked items."\\
\textbf{2. Entity Selection:} Select at least 5 such objects. They can be completely new independent entities or sub-parts of any entity that were not included in the [Sub-parts List]. ""The more items you can identify, the better, as we aim for the most comprehensive and detailed scene understanding possible."\\
\textbf{3. Contextual Description:} For each newly identified object, provide its name and a brief description of its appearance.\\
\textbf{4. Naming Rule:} Keep each `name` short and use common everyday words. ""Prefer one visible attribute when possible (not mandatory), from: `color / material / pattern / shape / size / state`. ""If you add an attribute, add only one. "\\
\textbf{Example:} use `small button` or `black button` instead of `small round metal button`.\\
Output pattern:\\
\textcolor{red}{\{output\_format\_text\}}\\
\textbf{Your output must be JSON only (no markdown)}.\\
\textbf{Data:}\\
scene graph:\textcolor{red}{\{graph\_text\}},\\
sub-part list:\textcolor{red}{\{sub\_part\_list\_text\}}.\\
\end{promptbox}
\caption{Prompt used for second-round global entity discovery.}
\label{tab:round2-prompt}
\end{table*}

New object names from both rounds are deduplicated before grounding. Each retained name is queried with GroundingDINO under the same $0.30$ box threshold and $0.25$ text threshold described above. Every retained box is segmented with SAM ViT-H.

\subsection{Two-stage Filtering}
\label{app:two-stage-filtering}

We merge masks from three sources. The first consists of masks for source entities. The second consists of additional masks obtained by grounding names in the source vocabulary. The last consists of masks generated from the vocabulary discovered by the VLM procedure. 
Filtering is necessary because open-vocabulary grounding can yield repeated masks for one name. It can also assign different names to the same region, or return a weak detection whose query does not match its segmented region.

\subsubsection{Geometric filtering.}
\textbf{Same Name Detection.}
We group candidates that share the same name. Within each group, let $M_i$ be the smaller mask and $M_j$ the larger one. We compute the fraction of $M_i$ covered by $M_j$ as
\begin{equation}
\operatorname{IoA}(i,j)=\frac{|M_i\cap M_j|}{|M_i|}.
\label{eq:ioa}
\end{equation}
If $\operatorname{IoA}(i,j)>0.95$, the masks are regarded as nested duplicates. We retain the larger mask only if its grounding score exceeds that of the smaller mask by more than $0.10$. Otherwise, we retain the smaller mask because it is more likely to preserve a fine-grained entity.

\textbf{Overlap Detection.}
We next compare each newly introduced candidate with the masks associated with entities in the source scene graph.
For a candidate mask $M_c$ and a source entity mask $M_s$, we compute their intersection over union (IoU).
If the IoU exceeds $0.90$, we remove $M_c$. 
At this level of overlap, the newly grounded region is treated as another description of an entity that has already been annotated in the source scene graph, rather than as an additional visual entity.

This rule gives priority to source annotations. A source mask is retained even when a newly grounded mask has a higher grounding score, because we assume that the entity annotations in the source scene graph are correct.

\subsubsection{Semantic filtering.}
Geometric filtering alone cannot decide whether two labels assigned to the same region are semantically interchangeable or genuinely different. We therefore use two local VLMs: Qwen2-VL-7B-Instruct~\citep{wang2024qwen2}and LLaVA-v1.6-Vicuna-13B~\citep{liu2024improved}.

\textbf{Candidate Name Disambiguation.}
After geometric filtering, each remaining candidate mask retains its associated entity name, which originates either from grounding names in the source scene graph or from the vocabulary proposed by the two VLM discovery rounds.
We compute the intersection over union (IoU) between each pair of candidate masks. 
We then apply complete linkage clustering with an IoU threshold of $0.9$, merging two clusters only when every pair of masks across them has an IoU greater than $0.9$. 
Consequently, all masks within each resulting cluster have pairwise IoU greater than $0.9$. 
We collect the distinct normalized names associated with the masks in each cluster into a candidate name list. 
We then overlay the union of the masks in each cluster on the original image and ask each VLM to select exactly one name from the corresponding list.
See Table~\ref{tab:candidate name selection} for our prompt.

\begin{table*}[!t]
\centering
\begin{promptbox}[Candidate name selection]
I will provide an ``Image" and an ``Object Name List".\\
\relax[Image]\\
- The image is a real-world scene.\\
- One target object has already been highlighted with a salient mask.\\
- You only need to identify the masked target object.\\
\relax[Object Name List]\\
The list contains candidate object names. Your task is to choose the single best-matching name for the masked target object.\\
\textbf{Instructions:}\\
1. Focus only on the masked region.\\
2. Compare the masked object with all candidate names.\\
3. Select the one candidate that best matches the masked target object.\\
4. Do not output any name outside the candidate list.\\
5. If several names are similar, choose the most specific visible one.\\
Output pattern:\\
{
``object\_ name": ``..." 
}\\
\textbf{Your output must be JSON only (no markdown)}\\
\textbf{Data:}\\
object name list:\textcolor{red}{\{object\_name\_list\_text\}}.
\end{promptbox}
\caption{Prompt used for candidate name selection.}
\label{tab:candidate name selection}
\end{table*}

Although geometric filtering removes clear duplicates, it cannot determine which of several names best describes the same region. 
We therefore use the two VLMs to make a constrained semantic decision before retaining an ambiguous candidate. 
When Qwen2-VL and LLaVA select the same candidate name, we keep the mask with the highest score under that name. 
If only one VLM returns a valid candidate name, or if the two VLMs disagree, we fall back to the grounding scores. We retain the highest-scoring mask in the cluster if its score exceeds $0.45$; otherwise, we discard the entire cluster.

\textbf{Low-confidence Object Verification.}
Semantic filtering also verifies whether the masks of low-confidence objects are correctly matched to their names.
We apply this check only to retained candidates outside the source scene graph, subject to two conditions: a grounding score below $0.37$ and a mask area exceeding $0.005$ of the image area. 
Both VLMs independently assess whether the current name matches the highlighted mask.
See Table~\ref{tab:prompt for verifying a name against a mask} for the prompt.

\begin{table*}
\centering
\begin{promptbox}[Prompt for verifying a name against a region]
I will provide an "Image", a "Target Object Name", and a "Decision List".\\
\relax[Image]\\
- The image is a real-world scene.\\
- One target object has already been highlighted with a salient mask.\\
- Focus only on the highlighted masked region. \\
\relax[Target Object Name]\\
- \textcolor{red}{\{object\_name\}}\\
\relax[Decision List]\\
- yes\\
- no\\
\textbf{Task:}Decide whether the masked target object matches the target object name.\\
\textbf{Your output must be JSON only (no markdown)}\\
\{object\_name: ``yes" or ``no"\}
\end{promptbox}
\caption{Prompt for verifying a name against a mask.}
\label{tab:prompt for verifying a name against a mask}
\end{table*}

If both models answer yes, the candidate is retained. 
Otherwise, we request a new object name from the VLM selected by the following rule. 
When one model answers yes and the other answers no, the model that answers no generates the replacement name. 
When both models answer no, Qwen2-VL-7B-Instruct is designated to generate the name. 
This rule assigns relabeling to the model that explicitly identifies a mismatch, and it provides a deterministic decision when both models reject the original label.

To verify that relabeling refers to the original visual region rather than to a semantically plausible object elsewhere in the image, the proposed name is marked as pending rather than accepted immediately. 
We rerun GroundingDINO and SAM using that name.
The new label is accepted only if at least one newly grounded mask has IoU greater than $0.90$ with the original mask.

\section{Detailed Implementation of Stage 2}
\subsection{Holistic Relation Construction}
\label{app:holistic_relation}

We provide the full image together with the region annotations of all entities retained after Stage~1 to the pretrained ROBIN-3B~\cite{park2025synthetic}, without additional fine-tuning.
These entities include both the original source entities and the newly discovered entities retained through Stage~1 filtering.
We set the generation temperature to $0.2$, top-$p$ to $1.0$, and the maximum output length to $1{,}024$ tokens.
The model returns directed relation predictions among these entities.

We first remove invalid predictions, including predictions that cannot be parsed or connect an entity to itself.
We then remove duplicate relations. Two predictions are considered duplicates if they have the same subject and object and their relation are identical. 
Since the source annotations already provide relations among source entities, we discard ROBIN predictions whose two endpoints both belong to the source entity set.
The retained relations therefore connect a newly discovered entity to either a source entity or another newly discovered entity.

\subsection{Dependency Relation Construction with the Dependency Candidate Tree}
\label{app:dependency_relation}
\textbf{Dependency Candidate Tree Construction.} For each entity $i$, we consider entity $j$ as a potential parent when its mask has a larger area. Let $M_i$ and $M_j$ denote their masks, with areas $A_i=|M_i|$ and $A_j=|M_j|$. We require $A_j>A_i$ and measure the fraction of $i$ covered by $j$ using
\begin{equation}
\mathrm{IoA}(i,j)=\frac{|M_i\cap M_j|}{|M_i|}.
\end{equation}

Dependency relations such as \emph{part of} and \emph{attached to} typically imply close physical association between the two entities, and thus the regions near their local boundary are expected to have similar relative depths. To assess local depth consistency, we use morphological dilation to extract narrow regions on the two sides of their local boundary.
Let $B_r$ denote an elliptical structuring element of size $(2r+1)\times(2r+1)$, and let $M \oplus B_r$ denote the morphological dilation of mask $M$.
We define
\begin{equation}
\begin{aligned}
R_{i\rightarrow j}
&=
\bigl((M_i \oplus B_r)\cap M_j\bigr)\setminus M_i,\\
R_{j\rightarrow i}
&=
\bigl((M_j\setminus M_i)\oplus B_r\bigr)\cap M_i.
\end{aligned}
\end{equation}
The first region $R_{i\rightarrow j}$ contains pixels within $M_j$ just outside $M_i$, while the second region $R_{j\rightarrow i}$ contains pixels within $M_i$ close to $M_j\setminus M_i$. 
Using the per-image normalized depth map, let $\bar{D}(R)$ denote the mean depth over valid pixels in region $R$.
If either region contains no valid depth values, the corresponding entity pair is excluded because $\Delta z(i,j)$ cannot be reliably computed.
Otherwise, we measure their local depth difference as
\begin{equation}
\Delta z(i,j)
=
\left|
\bar{D}(R_{i\rightarrow j})
-
\bar{D}(R_{j\rightarrow i})
\right|.
\end{equation}
We retain $j$ as a candidate parent of $i$ only when both the spatial overlap and local depth conditions are satisfied:
\begin{equation}
\mathrm{IoA}(i,j)>0.6,
\qquad
\Delta z(i,j)<0.035.
\end{equation}

Each retained pair defines a directed candidate edge $j\rightarrow i$, from parent to child. To select among competing candidate parents, we assign each edge the weight
\begin{equation}
s_{j\rightarrow i}
=
\lambda_1\bigl(\mathrm{IoA}(i,j)-0.8\bigr)
+
\lambda_2\left(1-\frac{\Delta z(i,j)}{0.02}\right)
-
\lambda_3\log\frac{A_j}{A_i},
\end{equation}
where $\lambda_1$, $\lambda_2$, and $\lambda_3$ are nonnegative weights controlling the contributions of mask coverage, depth consistency, and the area ratio, respectively. 
The weight $s_{j\rightarrow i}$ measures how suitable entity $j$ is as a candidate parent of entity $i$ in the Dependency Candidate Tree.
A larger score indicates stronger support for this assignment.
The first term rewards stronger containment of $i$ within $j$, and the second rewards similar local depths between the two entities.
The third term penalizes excessively large candidate parents, since a much larger region may contain $i$ without being its most relevant parent.

The edge weights between entities define a weighted directed graph in which an entity may have multiple possible parents.
We therefore need to select a single consistent parent assignment while maximizing the overall quality of the selected connections.
At the same time, some entities may not have a suitable parent and should remain independent.

To deal with such problems, we introduce a virtual root and connect it to every entity with a directed edge of weight zero.
We then apply the Chu--Liu/Edmonds algorithm~\citep{chu1965shortest,edmonds1967optimum} to obtain a maximum-weight spanning arborescence rooted at the virtual node.
This selects at most one parent for each entity while maximizing the total score of the selected edges.
Because the virtual-root edges have zero weight, an entity is attached to the virtual root when none of its candidate parents provides a positive score.
Finally, we remove the virtual root and its incident edges, leaving a forest that we refer to as the Dependency Candidate Tree.

\textbf{Dependency Relation Prediction.} 
The Dependency Candidate Tree determines which entity pairs are considered for dependency relation prediction.
For each selected tree edge $j\rightarrow i$, we first examine the provenance of the two entities.
If both entities originate from the source scene graph, we skip relation prediction for this pair because their relations are already provided by the source annotations.
We therefore apply dependency relation prediction only to edges involving at least one newly discovered entity.

For each retained edge, we use Qwen2-VL-7B-Instruct~\citep{wang2024qwen2} to infer the dependency relation between the two entities.
The model receives the image together with the names of entities $j$ and $i$, where $j$ and $i$ are specified as the candidate parent and child in the Dependency Candidate Tree, respectively.
The model is instructed to predict a single dependency predicate for the entity pair. 
The predictions are combined with the relations retained from the source scene graph in the subsequent relation integration step.

\section{Detailed Implementation of FiGRA}
\subsection{Two-stage Training}
\label{app:two_stage_training}

\subsubsection{Pre-training}
We pretrain the graph-side modules to learn multimodal graph representations and map them to continuous tokens for a frozen MLLM. Training uses three objectives.

\textbf{Shortest-Path Prediction.}
The MLLM predicts the shortest path between entities $(s,t)$ as a triplet sequence $Y^{\mathrm{sp}}$ (e.g., "$(s,\text{is on},b),(b,\text{on the left of},t)$", where $b$ denotes an intermediate entity.)
We use the projected representations of the source node $\tilde{\mathbf{h}}_s$ and target node $\tilde{\mathbf{h}}_t$ as endpoint tokens to specify the starting and ending nodes of the path.
The prediction loss is
\begin{equation}
\begin{aligned}
\mathcal{L}_{\mathrm{sp}}
=-\frac{1}{|Y^{\mathrm{sp}}|}
\sum_k \log p_{\theta_0}\!\bigl(
y_k^{\mathrm{sp}}\mid y_{<k}^{\mathrm{sp}},
\mathbf{T}_I,\mathbf{T}_g,
\tilde{\mathbf{h}}_s,\tilde{\mathbf{h}}_t,Q^{\mathrm{sp}}
\bigr),
\end{aligned}
\end{equation}
where $\theta_0$ denotes the frozen MLLM parameters and $Q^{\mathrm{sp}}$ is the path prediction instruction (See~\cref{app:instructions} for detailed instruction).

\textbf{Node Feature Reconstruction.}
To preserve the textual and visual information in the node representations, we use two linear decoders to reconstruct the original features of both modalities from the node representations $\mathbf{H}$:
\begin{equation}
\mathcal{L}_{\mathrm{feat}}
=\frac{1}{2}\left[
\operatorname{MSE}\!\left(D_t(\mathbf{H}),\mathbf{X}^{text}\right)
+\operatorname{MSE}\!\left(D_v(\mathbf{H}),\mathbf{X}^{vision}\right)
\right].
\end{equation}
where $\mathbf{H}=[\mathbf{h}_v^{(L)}]_{v\in\mathcal{V}}$ denotes the node representations, $\mathbf{X}^{text}=[\mathbf{x}_v^{text}]_{v\in\mathcal{V}}$ and $\mathbf{X}^{vision}=[\mathbf{x}_v^{vision}]_{v\in\mathcal{V}}$ contain the original text and visual features, respectively, and $D_t$ and $D_v$ denote the corresponding linear decoders.

\textbf{Topology Reconstruction.}
We predict whether an ordered node pair is connected by a directed edge using
\begin{equation}
\ell_{uv}^{(i)}=D_{\mathrm{topo}}([\mathbf{h}_u^{(L,i)};\mathbf{h}_v^{(L,i)}]),\qquad
\hat{a}_{uv}^{(i)}=\sigma(\ell_{uv}^{(i)}),
\end{equation}
where $D_{\mathrm{topo}}$ is an MLP and $\ell_{uv}^{(i)}$ denotes the predicted logit for the node pair $(u,v)$ in graph $i$. For each graph, we construct a set of positive and sampled negative pairs $\mathcal{P}_i$. The topology reconstruction loss is
\begin{equation}
\mathcal{L}_{\mathrm{topo}}
=-\frac{1}{B}\sum_{i=1}^{B}\frac{1}{|\mathcal{P}_i|}
\sum_{(u,v)\in\mathcal{P}_i}
\left[
a_{uv}^{(i)}\log\hat{a}_{uv}^{(i)}
+(1-a_{uv}^{(i)})\log(1-\hat{a}_{uv}^{(i)})
\right].
\end{equation}
where $B$ is the batch size, and $a_{uv}^{(i)}$ indicates whether the directed edge from node $u$ to node $v$ exists in graph $i$.

\subsubsection{Supervised Fine-tuning}
We load the pretrained parameters for the multimodal fusion, graph encoder, and adapter modules, and adapt FiGRA to downstream tasks by jointly training these modules and LoRA parameters, while keeping the original MLLM parameters frozen. Given a task instruction $Q$ and a reference answer $Y$, we minimize the causal language modeling loss:
\begin{equation}
\mathcal{L}_{\mathrm{SFT}}
=-\frac{1}{|Y|}\sum_k
\log p_{\theta}\!\left(
y_k\mid y_{<k},\mathbf{T}_I,\mathbf{T}_g,Q
\right).
\end{equation}
We use this objective alone during SFT, with all three pre-training objectives disabled.

\subsection{Instructions}
\label{app:instructions}
Please see~\cref{tab:sft_prompt} for our instructions $Q$ used in supervised fine-tuning stage.
Please see~\cref{tab:pre-training-prompt} for our instructions used in pre-training stage.

\begin{table*}[t]
\centering
\begin{orangepromptbox}[FiGRA Fine-tuning Prompt]
\textbf{SYSTEM:}\\
You are a knowledgeable image description generating agent.\\
Your task is to generate detailed image descriptions using the input image and graph semantic tokens.\\
Graph semantic tokens are distilled from the multimodal scene graph associated with the image and contain rich information about detailed scene context.\\
\textbf{USER:}\\
\textcolor{red}{[Image]}\\
Question: \textcolor{red}{[Question]}\\
Here are graph semantic tokens: \textcolor{red}{[Graph Semantic Tokens]}\\
\textbf{ASSISTANT:}
\end{orangepromptbox}
\caption{Prompt used for supervised fine-tuning of FiGRA.}
\label{tab:sft_prompt}
\end{table*}

\begin{table*}[t]
\centering
\begin{orangepromptbox}[FiGRA Pre-training Prompt]
\textbf{SYSTEM:}\\
You are a graph reasoning assistant. Predict the shortest path from the source node to the target node using the image and graph representations.\\
\textbf{USER:}\\
\textcolor{red}{[Image]}\\
Question: Predict the shortest path from the source node to the target node.\\
Source node representation: \textcolor{red}{[Source Node Token]}\\
Target node representation: \textcolor{red}{[Target Node Token]}\\
Here are graph semantic tokens: \textcolor{red}{[Graph Semantic Tokens]}\\
\textbf{ASSISTANT:}
\end{orangepromptbox}
\caption{Prompt used for shortest-path prediction during pre-training.}
\label{tab:pre-training-prompt}
\end{table*}

\subsection{Hyperparameter Settings}
The hyperparameter settings for pre-training and supervised fine-tuning are summarized in Tables~\ref{tab:pretraining_hyperparameters} and~\ref{tab:finetuning_hyperparameters}, respectively.
\begin{table}[t]
\centering
\caption{Key hyperparameters for graph pre-training.}
\label{tab:pretraining_hyperparameters}
\small
\renewcommand{\arraystretch}{1.08}
\begin{tabular}{@{}>{\centering\arraybackslash}p{0.56\columnwidth}>{\centering\arraybackslash}p{0.20\columnwidth}@{}}
\toprule
Hyperparameter & Value \\
\midrule
\rowcolor{gray!20}
\multicolumn{2}{c}{\textit{\textbf{Model Architecture}}} \\
Graph hidden dimension ($d$) & 768 \\
MLLM token embedding dimension ($d_{MLLM}$) & 4096 \\
Graph Transformer layers ($L$) & 2 \\
Attention heads per layer & 8 \\
Graph dropout & 0.1 \\
\midrule
\rowcolor{gray!20}
\multicolumn{2}{c}{\textit{\textbf{Training Objectives}}} \\
Shortest-path loss weight ($\lambda_{\mathrm{sp}}$) & 1.0 \\
Feature reconstruction loss weight ($\lambda_{\mathrm{feat}}$) & 0.3 \\
Topology reconstruction loss weight ($\lambda_{\mathrm{topo}}$) & 0.1 \\
\midrule
\rowcolor{gray!20}
\multicolumn{2}{c}{\textit{\textbf{Optimization}}} \\
Optimizer & AdamW \\
Learning rate & $1\times10^{-4}$ \\
Weight decay & $1\times10^{-2}$ \\
Learning-rate schedule & Cosine decay \\
Warmup ratio & 0.05 \\
Batch size & 1 \\
Gradient accumulation steps & 8 \\
Training epochs & 5 \\
\bottomrule
\end{tabular}
\end{table}

\begin{table}[t]
\centering
\caption{Key hyperparameters for supervised fine-tuning. Triple retrieval is applied to FG-Caption and FG-QA.}
\label{tab:finetuning_hyperparameters}
\small
\renewcommand{\arraystretch}{1.08}
\begin{tabular}{@{}>{\centering\arraybackslash}p{0.56\columnwidth}>{\centering\arraybackslash}p{0.20\columnwidth}@{}}
\toprule
Hyperparameter & Value \\
\midrule
\rowcolor{gray!20}
\multicolumn{2}{c}{\textit{\textbf{Model Architecture}}} \\
Graph hidden dimension ($d$) & 768 \\
MLLM token embedding dimension ($d_{MLLM}$) & 4096 \\
Graph Transformer layers ($L$) & 2 \\
Attention heads per layer & 8 \\
Graph dropout & 0.1 \\
\midrule
\rowcolor{gray!20}
\multicolumn{2}{c}{\textit{\textbf{LoRA Configuration}}} \\
LoRA rank & 8 \\
LoRA scaling parameter ($\alpha$) & 16 \\
LoRA dropout & 0.05 \\
\midrule
\rowcolor{gray!20}
\multicolumn{2}{c}{\textit{\textbf{Graph Retrieval}}} \\
Retrieved triples per sample ($K$) & 7 \\
Graph expansion hops & 1 \\
\midrule
\rowcolor{gray!20}
\multicolumn{2}{c}{\textit{\textbf{Optimization}}} \\
Optimizer & AdamW \\
Graph module learning rate & $1.5\times10^{-4}$ \\
LoRA learning rate & $1\times10^{-5}$ \\
Weight decay & $3\times10^{-3}$ \\
Learning-rate schedule & Cosine decay \\
Warmup ratio & 0.06 \\
Batch size & 1 \\
Gradient accumulation steps & 8 \\
Training epochs & 5 \\
\bottomrule
\end{tabular}
\end{table}

\section{Baselines}
\label{app:baselines}
\textbf{Open-source MLLMs.}
Our open-source multimodal large language model baselines consist of LLaVA-1.5-7B~\cite{liu2024improved}, Qwen2-VL-7B-Instruct~\cite{wang2024qwen2}, ALLaVA-3B~\cite{chen2024allava}, Llama-3.2-11B-Vision-Instruct~\cite{grattafiori2024llama}, and InternVL3.5-8B~\cite{wang2025internvl3}.
We freeze the vision encoder and the base language model, and apply LoRA only to the attention projection layers of the language model. 
All models are trained using autoregressive cross-entropy loss on the ground-truth responses.

\subsection{FG-QA Baselines}
\textbf{ViCrop~\cite{khayatkhoei2025mllms}.}
ViCrop is a training-free approach for fine-grained visual question answering.
It uses attention maps or input gradients from a frozen MLLM to estimate image region importance, selects a crop by maximizing the importance score within sliding windows at multiple scales, and adds the resized crop to the original image for a second prediction. 
We use ViCrop as a training-free baseline without fine-tuning the backbone MLLM.

\textbf{PinPoint~\cite{kwon2026focus}.}
PinPoint trains a region selector using annotated image regions as supervision. 
The original method uses bounding boxes to annotate answer regions and supporting evidence, with an encompassing region covering both. 
The selector is trained on the training split of each dataset and evaluated on its test split.

To adapt PinPoint to FG-QA, we construct encompass boxes from our scene graphs. 
For each question, the encompass box is the smallest rectangle enclosing the bounding boxes of the entities mentioned in the question and the entity corresponding to the answer. 
We use these encompass boxes to supervise the training of separate PinPoint selectors with LLaVA-1.5-7B and Qwen2-VL-7B-Instruct as backbones, and evaluate both models on the FG-QA test split.

\textbf{SEAL~\cite{wu2024v}.}
SEAL is a multimodal visual question answering framework that combines a VQA LLM with a Visual Search Model (VSM). We use it without additional fine-tuning.
The VQA model first identifies visual details needed to answer the question but missing from the full image. 
The VSM then recursively searches the image for these targets and returns the corresponding crops and their locations to the VQA model, which uses this additional visual information to produce the answer.

\textbf{Zoom Eye~\cite{shen2025zoomeye}}
Zoom Eye also starts from the full image. When that view is insufficient, it zooms progressively into candidate regions. The model can return to an earlier view and follow a different branch when needed.
We use Zoom Eye without additional fine-tuning. At inference time, the MLLM identifies question-relevant visual cues and uses confidence-based tree search to explore candidate regions. 
It then answers from the visual context gathered during the search.

\subsection{FG-Description and FG-Caption Baselines}
\textbf{SC-Captioner~\cite{zhang2025sc}.}
SC-Captioner is an image-captioning model based on Qwen2-VL. 
Its training proceeds in two stages. 
The first freezes the base model and trains a LoRA adapter to generate reference captions. 
This adapter is subsequently merged into the base model.
The second stage trains the model to correct its own captions through reinforcement learning. 
Given a fixed correction instruction, the model revises an initial caption. 
Both captions and the reference are parsed into objects, attributes, and relations. 
The reward favors revisions that remove unsupported facts or add reference-consistent ones. It penalizes revisions that delete correct information, introduce inconsistencies, or leave the caption unchanged. KL regularization against the frozen SFT model stabilizes the process. We use the revised caption as the final output.

\textbf{ShareGPT4V-7B~\cite{chen2024sharegpt4v}.}
ShareGPT4V-7B is a multimodal large language model built upon the LLaVA-1.5 architecture, consisting of a CLIP-Large vision encoder, a two-layer MLP projector, and a Vicuna-v1.5-7B language model. 
It leverages high-quality image captions from ShareGPT4V to improve vision-language alignment and detailed visual understanding. 
In our experiments, we use the ShareGPT4V-7B as an image captioning baseline and evaluate it both without additional fine-tuning and after applying LoRA to the language model's attention projection layers.

\textbf{IT~\cite{pi2024image}.}
IT is a training-free image textualization framework that combines a pretrained MLLM with vision expert models to generate detailed image descriptions.
It first generates a holistic reference description and extracts object entities from it.
Vision expert models are then employed to detect hallucinated objects and obtain fine-grained visual information, including object attributes, spatial positions, depths, and sizes.
Finally, an LLM uses the reference description and the extracted visual information to generate a refined caption.
In our implementation, we use a pretrained LLaVA model as the backbone MLLM for IT and adopt the training-free setting.

\textbf{Patch Matters~\cite{peng2025matters}.}
Patch Matters is a training-free captioning method that combines global and regional visual observations.
It first constructs a global view and multiple candidate image patches, and uses a pretrained multimodal language model to describe the complete image and the selected regions.
The global caption, regional captions, and spatial information of the patches are subsequently combined and provided to a language model, which produces the final detailed caption.
We adopt the training-free setting and do not fine-tune the model on FG-Caption.

\textbf{Synthesized~\cite{dong2024benchmarking}.}
Synthesized is a five-stage detailed caption construction pipeline.
It first uses a pretrained LVLM to generate an overall image caption, then applies SAM to identify salient regions and generates local captions for them using the same LVLM.
A phrase-level hallucination filtering stage uses the Factual parser to extract visual elements and OWLv2 to remove unsupported phrases.
Finally, an LLM merges the filtered overall and local captions into a coherent, detailed description.
We run the complete pipeline at inference time, without fine-tuning its component models.

\section{Case Study}
\label{app:case_study}

\begin{figure}[t]
\centering
\includegraphics[
    width=0.85\linewidth,
    height=0.78\textheight,
    keepaspectratio
]{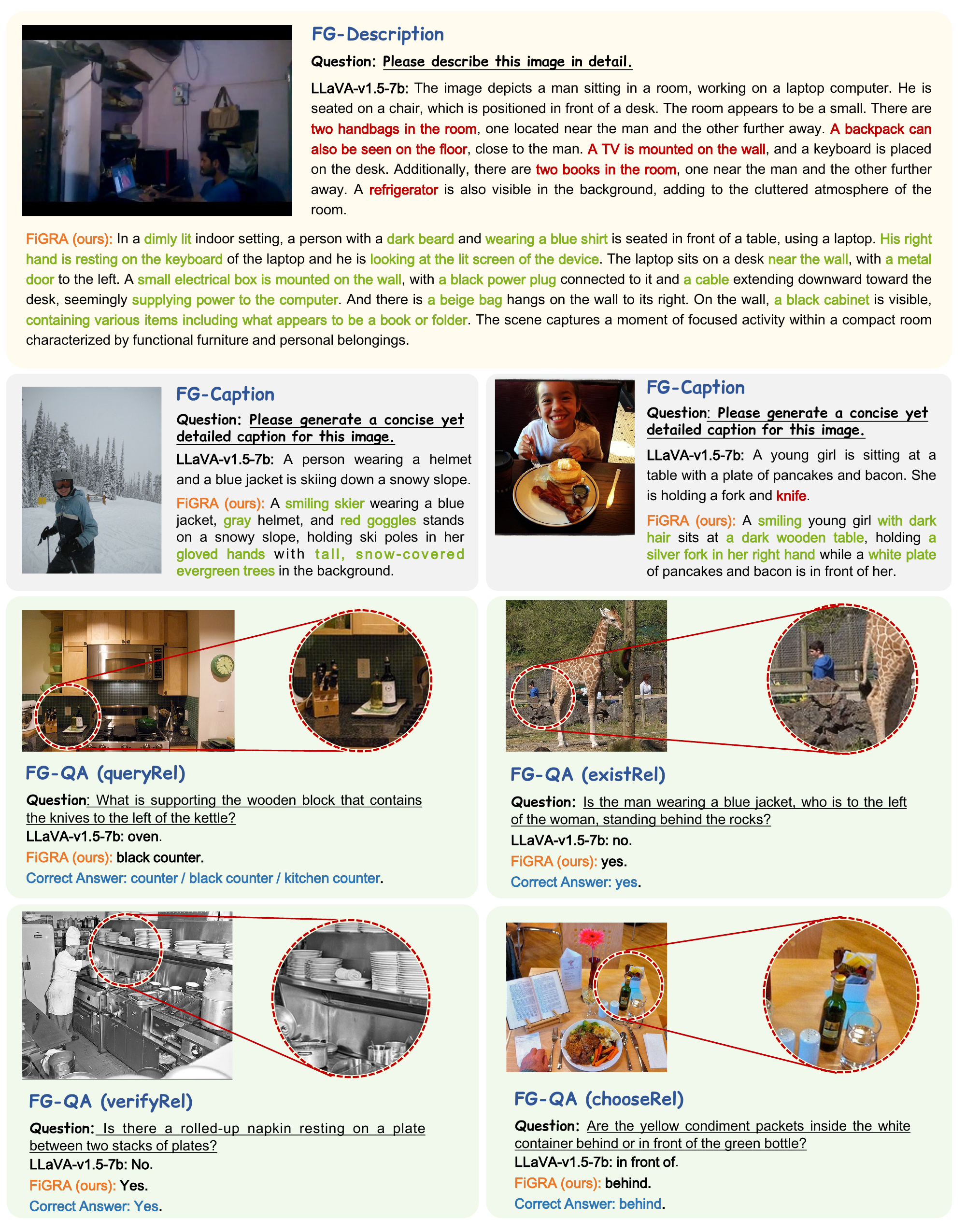}
\caption{
Case Study of FiGRA and LLaVA-1.5-7B on FG-GraphBench. 
Green text highlights fine-grained details captured by FiGRA that are missing from LLaVA's answers.
Red text indicates incorrect details generated by the model.
}
\label{fig:case_study}
\end{figure}

Figure~\ref{fig:case_study} presents a qualitative comparison between FiGRA and LLaVA-1.5-7B on FG-GraphBench. 
Compared with LLaVA-1.5-7B, FiGRA captures and describes more fine-grained visual details, including information that is either omitted or described less precisely in the LLaVA's responses. 
These differences are highlighted in green. 
Meanwhile, red text indicates incorrect details in the generated answers.
This comparison illustrates the potential of incorporating fine-grained scene graph information to improve MLLMs' ability to perceive fine-grained visual details and generate accurate responses.

\end{document}